\documentclass{article}
\usepackage[T1]{fontenc} % Obvs needed
\usepackage[utf8]{inputenc} % Obvs needed
\usepackage{lmodern} % For guillemets
\usepackage{amsmath,amsfonts,amssymb,amsthm,mathrsfs} % For math equations, theorems, symbols, etc
\usepackage{xcolor} % For naming colors
\usepackage{array} % Creating arrays
\usepackage{enumitem} % For enumerating
\usepackage{graphicx}
\usepackage{float}
\usepackage{dsfont}
\usepackage{fancyhdr}
\usepackage{graphicx}
\usepackage{caption}
\usepackage{xcolor}
\usepackage{booktabs} % for professional tables

\usepackage{parskip}

\usepackage{empheq}
\usepackage{rotating}

\usepackage[margin=1.in]{geometry}

\usepackage{algorithmic}
\usepackage{algorithm}

\theoremstyle{plain}
\newtheorem{theorem}{Theorem}[section]

\newtheorem{property}{Property}

\theoremstyle{definition}
\newtheorem{definition}[theorem]{Definition}

\theoremstyle{remark}

\usepackage{amsmath}
\usepackage{amssymb}
\usepackage{mathtools}
\usepackage{amsthm}

\usepackage{microtype}
\usepackage{graphicx}
\usepackage{subcaption}
\usepackage{booktabs} % for professional tables

\usepackage{natbib}
\usepackage{amsmath,amsfonts,bm}

\def\eqref#1{equation~\ref{#1}}
\def\1{\bm{1}}

\def\rvd{{\mathbf{d}}}

\def\vtheta{{\bm{\theta}}}

\def\vv{{\bm{v}}}
\def\vw{{\bm{w}}}
\def\vx{{\bm{x}}}
\def\vy{{\bm{y}}}
\def\vz{{\bm{z}}}

\DeclareMathAlphabet{\mathsfit}{\encodingdefault}{\sfdefault}{m}{sl}
\SetMathAlphabet{\mathsfit}{bold}{\encodingdefault}{\sfdefault}{bx}{n}

\def\sT{{\mathbb{T}}}

\def\sX{{\mathbb{X}}}
\def\sY{{\mathbb{Y}}}

\newcommand{\normltwo}{L^2}

\DeclareMathOperator*{\argmin}{arg\,min}

\usepackage{hyperref}
\usepackage{authblk}
\title{Managing Solution Stability in Decision-Focused Learning with Cost Regularization}
\author[1,2]{Victor Spitzer}
\author[1]{Francois Sanson}
\affil[1]{Lhyfe, Nantes, France}
\affil[2]{LISN, Université Paris-Saclay, Gif-sur-Yvettes, France}

\begin{document}

\maketitle

\begin{abstract}
  Decision-focused learning integrates predictive modeling and combinatorial optimization by training models to directly improve decision quality rather than prediction accuracy alone. Differentiating through combinatorial optimization problems represents a central challenge, and recent approaches tackle this difficulty by introducing perturbation-based approximations. In this work, we focus on estimating the objective function coefficients of a combinatorial optimization problem. Our study demonstrates that fluctuations in perturbation intensity occurring during the learning phase can lead to ineffective training, by establishing a theoretical link to the notion of solution stability in combinatorial optimization. We propose addressing this issue by introducing a regularization of the estimated cost vectors which improves the robustness and reliability of the learning process, as demonstrated by extensive numerical experiments.
\end{abstract}

Recent advances  in data processing and analytics have led to a shift in operations management for supply chains, healthcare systems, and industries \citep{Mivsic_2020}. Data is integrated into decision-making processes by training Machine Learning (ML) models to create mappings between observed features and the parameters of an optimization problem \citep{Qi_2022}. In most situations, prediction errors from ML models result in suboptimal decisions and unnecessary costs that could have been prevented \citep{Cameron_2021}. A growing body of research is advancing the \textit{Decision-Focused Learning} (DFL) paradigm, which tackles this challenge by embedding combinatorial optimization into the learning process to enhance prediction quality for specific optimization tasks.

We focus in this work on estimating costs in the objective function of combinatorial optimization problems taking the form of \textit{Mixed-Integer Linear Programs} (MILP). These problems are modeled by continuous and discrete variables, with a linear objective function to optimize and a set of linear constraints to respect. For this specific task, decision-focused learning most often outperforms standard ML models, but it is hindered by the discontinuity found in decision models mappings. Nevertheless, a large variety of approximate techniques has been proposed in recent years to circumvent this difficulty and learn from discrete decision models, most of them relying on decisions perturbations to enable differentiation. In practice, managing the intensity of these perturbations can be challenging and may result in underwhelming performance.

Our primary contribution is to expose a limitation shared by all examined perturbation-based DFL techniques, grounded in key principles of optimal solution stability in combinatorial optimization. Specifically, we demonstrate that, depending on the perturbation intensity relative to the estimated costs, certain methods may resort to imitating known solutions rather than improving a target performance metric, while others may become entirely ineffective. Whereas previous work often overlooks perturbation intensity-leading to degraded learning behavior-our study shows that regularizing the estimated costs provides better control over the decision-focused learning process, a finding supported by extensive numerical experiments on multiple established benchmarks.

\section{Related work}

The integration of prediction and optimization in ML models has gained significant attention in recent years for its potential to improve decision-making under uncertainty \citep{Kotary_2021, Mandi_2024}. \citet{Mandi_2024} outline several categories of DFL techniques dedicated to the prediction of an optimization problem objective. A first category analytically differentiates certain optimization mappings via their Lagrangian formulations \citep{Amos_2017}. While powerful in theory, this approach requires additional smoothing for linear programs and the relaxation of integrality constraints for mixed-integer problems, introducing substantial approximations when applied to MILPs \citep{Wilder_2019}. Because our focus is on perturbation-based methods for MILP optimization, we do not further consider this line of work.

Another category relies on constructing smooth approximations of optimization mappings by the introduction of random perturbations. Originating from implicit differentiation via parameter perturbations \citep{Domke_2010}, this idea was adapted for DFL by considering distributions over decisions rather than single optimal solutions, enabling differentiation through discrete mappings. This adaptation produced the \textit{Perturb-and-MAP} framework (MAP) \citep{Papandreou_2011}, later refined in Implicit Maximum Likelihood Estimation (IMLE) \citep{Niepert_2021}, which improved perturbation sampling. The \textit{Differentiable Black-Box} (DBB) approach \citep{Pogancic_2020} built on similar principles but employed different perturbation strategies, while \textit{Differentiable Perturbed Optimizers} (DPO) \citep{Berthet_2020} extended Perturb-and-MAP by treating the optimization mapping as a deterministic function of both cost estimates and perturbations.

The last category gathers techniques with a dedicated loss function rather than a general differentiation framework. We take an interest in two state-of-the-art loss functions among the existing approaches. First, the \textit{Smart Predict-the-Optimize} (SPO) loss \citep{Elmachtoub_2022} corresponds to a surrogate regret loss function whose sub-gradients indicates worthwhile descent directions. We show that it relies on the introduction of a perturbation for optimization mapping differentiation, hence the need for regularization. Secondly, the \textit{Fenchel-Young} loss \citep{Blondel_2020} is a differentiable imitation loss function designed to reproduce the ground truth optimal decision. We compare the previously mentioned approaches to this loss function to analyze when learning behaviors shift from performance-driven improvement to simple imitation.

From an optimization perspective, \citet{Bengio_2021} introduce a distinction broadly categorizing DFL techniques as learning either by imitation of expert behavior, or by experience aimed at improving performance. We adopt this classification to demonstrate that perturbation-based DFL techniques designed for learning by experience may become inefficient-or collapse into imitation-if the cost vector estimates are not properly regularized. To support this claim, we draw on classical results related to solution stability in combinatorial optimization \citep{Bonnans_2000}.

While some studies mention regularization of cost vector estimates, it is typically introduced as a problem-specific adjustment-for instance, to address particular DFL applications \citep{Rolinek_2020} or as part of an extension of the DBB approach \citep{Sahoo_2022}. \citet{Blondel_2020} also mention regularization, although it is of a different nature as it applies to the optimization mapping for differentiation rather than to the cost vector itself. To date, regularization has not been recognized as a broadly applicable practice, independent of the chosen DFL method or the underlying optimization problem.

\section{General Framework}

Consider a convex cone $\Theta \subset\mathbb{R}^n$ and a non-empty finite set of distinct points $\sY \subset \mathbb{R}^n$ with $\mathcal{C}$ its convex hull. We consider a general MILP with linear costs parameterized by a vector $\vtheta \in \Theta$ and an associated \textit{optimization mapping} $f:\Theta \mapsto \mathcal{C}$ defined as follows:
\begin{equation}
\label{eq:optim}
\min_{y \in \mathcal{C}} \vtheta^\intercal y, \quad f(\vtheta)= \argmin_{y \in \mathcal{C}} \vtheta^\intercal y
\end{equation}
Function $f$ maps a cost vector $\vtheta \in \Theta$ to an optimal decision $f(\vtheta) \in \mathcal{C}$. Our goal is to estimate a cost vector $\vtheta$ whose quality depends on the resulting decision $f(\vtheta)$.

\subsection{Learning from Decisions}
We design a machine-learning model parameterized by $\vv$ and described by equations:
\begin{equation}
\label{eq:ML_model}
\vtheta = h_{\vv}(\vx), \quad \vy=f(\vtheta)
\end{equation}
Vector $\vx \in \sX$ denotes feature inputs while $\vy \in \sY$  is a target output representing a solution to a discrete decision model parameterized by $\vtheta \in \Theta$. Function $h_{\vv} : \sX \mapsto \Theta$ is a differentiable parameterized mapping while $f: \Theta \mapsto \sY$ represents an optimization mapping as described above. Here, the optimization mapping can be seen as a final combinatorial optimization layer within our ML model. We consider a training loss function based on an evaluation $l(\vy,\bar{\vtheta})$ of the estimated decision $\vy = f(h_{\vv}(\vx))$ with respect to the ground-truth cost vector $\bar{\vtheta}$. The loss is defined as:
\begin{equation}
\label{eq:loss}
L(\vx,\bar{\vtheta};\vv) = l(\vy,\bar{\vtheta}), \quad \text{ with } \vy=f(h_{\vv}(\vx))
\end{equation}
This DFL approach aims to predict the cost objective associated with the optimization mapping. Given a dataset $\{(\vx_j,\bar{\vtheta}_j)\}_{j=1}^N$, we learn parameters $\vv$ of \eqref{eq:ML_model} to minimize the average loss $L(\vx_j,\bar{\vtheta}_j;\vv)$ across all samples. In other words, we aim at learning a model that minimizes the evaluation value of decisions $\vy=f(\vtheta)$ made based on cost vector estimates $\vtheta=h_{\vv}(\vx)$.

This is typically observed when training a machine learning model using the \textit{regret} loss \( L^r \), which quantifies how much worse the decision based on predicted parameters is compared to the decision based on the true parameters. Specifically, it evaluates the decision estimate \( f(\vtheta) \in \mathcal{C} \subset \mathbb{R}^n \) with respect to the ground-truth cost vector \( \bar{\vtheta} \in \Theta \subset \mathbb{R}^n \).
\begin{equation} \label{eq:regret_loss}
    L^r(\vx,\bar{\vtheta};\vv) = \bar{\vtheta}^\intercal \vy  - \bar{\vtheta}^\intercal f(\bar{\vtheta}), \quad \text{ with } \vy=f(h_{\vv}(\vx))
\end{equation}
Although our examples and numerical experiments focus on this specific loss function and model structure, our contributions generalize to all studied perturbation-based DFL techniques within the broader framework introduced in~\eqref{eq:ML_model}.

Differentiating any training loss function $L$ as defined in \eqref{eq:loss} with regards to $\vv$ is necessary in order to train our model. According to \eqref{eq:loss}, this gradient can be expressed as:
\begin{equation}
\label{eq:L_gradient}
    \nabla_{\vv} L(\vx,\bar{\vtheta};\vv)=  \nabla_{\vv} h_{\vv}(\vx)^\intercal\nabla_{\vtheta} L(\vx,\bar{\vtheta};\vv)
\end{equation}
Hence it is necessary to differentiate the training loss function with regards to $\vtheta$. By chain rule this particular gradient can be expressed as follows:
\begin{equation}
\label{eq:gradient_loss}
\nabla_{\vtheta} L(\vx,\bar{\vtheta};\vv) = \nabla_{\vy}  L(\vx,\bar{\vtheta};\vv)^\intercal \nabla_{\vtheta} f(\vtheta)
\end{equation}
We conclude that in the case of the regret loss $L^r$, one has:
\begin{align}
    &\nabla_{\vy} L^r(\vx,\bar{\vtheta};\vv) = \bar{\vtheta} \label{eq:regret_diff_y} \\
    &\nabla_{\vtheta} L^r(\vx,\bar{\vtheta};\vv) = \bar{\vtheta}^\intercal \nabla_{\vtheta} f(\vtheta) \label{eq:regret_diff_theta}
\end{align}
Differentiating a loss function in a decision-focused learning model is challenging mainly because \(\nabla_{\boldsymbol{\theta}} L^r(\vx,\bar{\vtheta};\vv)\) depends on \(\nabla_{\boldsymbol{\theta}} f(\boldsymbol{\theta})\). The optimization mapping \(f\) is typically assumed to be piecewise constant, so the gradient \(\nabla_{\boldsymbol{\theta}} f(\boldsymbol{\theta})\) is zero almost everywhere. Since the gradient provides little value for optimization, DFL methods are applied to find effective descent directions for minimizing the loss function.

This is typically achieved by introducing cost perturbations within the optimization mapping, hence the prevalence of perturbation-based techniques among state-of-the-art DFL approaches. Indeed, among all existing methods, the only ones capable of learning the cost objective of mixed-integer linear problems rely on the introduction of perturbations.

\subsection{Stability Properties of an Optimization Mapping}

It is commonly assumed that the optimization mapping \( f \) is piecewise constant and thus possesses a gradient that is of little relevance in an optimization process \citep{Mandi_2024}. We strengthen this claim by reframing it into a stronger formalism based on combinatorial optimization literature related to optimal solution stability \citep{Bonnans_2000}. Moreover, we highlight how solution stability can characterize the impact of additive cost perturbations on the optimization mapping. The presented results are illustrated by an example in section \ref{section:solution_stability}.
 
Let us denote by $F$ a set valued optimization mapping such that $F(\vtheta)$ represents the exhaustive set of optimal solutions for the problem defined by \eqref{eq:optim} with cost vector $\vtheta \in \Theta$. In other words, $F(\vtheta)$ represents the set of all possible values taken by $f(\vtheta)$. We study the concept of upper semi-continuity for the set valued optimization mapping $F$ as presented by \citet{Bonnans_2000}.
\begin{definition}
    The set valued optimization mapping $F$ is upper semi-continuous at $\vtheta$ if for every open set $\mathcal{U}$ containing $F(\vtheta)$, there exists a neighborhood $\mathcal{V}$ of $\vtheta$ such that for all $\Tilde{\vtheta} \in \mathcal{V}$, one has $F(\Tilde{\vtheta}) \subset \mathcal{U}$.
\end{definition}
\citet{Bonnans_2000} identify sufficient conditions for a set valued mapping to be upper semi-continuous. These conditions are met in our case, given that polytope $\mathcal{C}$ is non-empty and bounded as the convex hull of the non-empty finite set $\mathbb{Y}$. Hence the following property:
\begin{property} \label{prop:F_usc}
    Let us consider the set valued optimization mapping $F$ for the problem defined by \eqref{eq:optim}. The mapping $F$ is upper semi-continuous everywhere in $\Theta$.
\end{property}
In most cases, the optimal solution set $F(\vtheta)$ consists only of a singleton $f(\vtheta)$ \citep{Elmachtoub_2022}. We restrict our study to that general case in what follows. In that case according to property \ref{prop:F_usc}, there exists a largest positive scalar $\eta$ such that every perturbed cost vector $\tilde{\vtheta}\in\Theta$ lying within distance $\eta$ of $\vtheta$ shares the same optimal decision set as $\vtheta$, that is, $F(\tilde{\vtheta}) = F(\vtheta)$. Borrowing from the nomenclature related to stability analysis, the maximum distance $\eta$ is defined as the \textit{stability radius} of cost vector $\vtheta$, and the neighborhood it defines around $\vtheta$ is its \textit{stability region} \citep{Sotskov_1995}. We conclude that, in most cases, the set valued optimization mapping $F$ is constant within a stability region around $\vtheta$, depending on a stability radius $\eta$. Hence the following property holds.

\begin{property} \label{prop:perturbation_stability}
    Consider an optimization mapping $f:\Theta \mapsto \mathcal{C}$ and a cost vector $\vtheta$ such that its optimal decision set $F(\vtheta)$ is a singleton $f(\vtheta)$. Let us further consider a perturbation $\bm{\delta} \in \Theta$ of the cost vector $\vtheta$ such that its optimal decision set $F(\bm{\delta})$ is a singleton $f(\bm{\delta})$. One has:
    \begin{itemize}
        \item If $\vtheta + \bm{\delta}$ in the stability region of $\vtheta$ then $f(\vtheta + \bm{\delta})=f(\vtheta)$
        \item If $\vtheta + \bm{\delta}$ in the stability region of $\bm{\delta}$ then $f(\vtheta + \bm{\delta})=f(\bm{\delta})$
    \end{itemize}
\end{property}

This stability property in face of additive perturbation implies that the optimization mapping at $\vtheta \in \Theta$ remains constant for any perturbation $\bm{\delta}$ within its stability region. This supports the claim that the optimization mapping is generally piecewise constant which explains the little relevance of its gradient in an optimization process. Moreover, property \ref{prop:perturbation_stability} represents the main motivation to an efficient management of solution stability in decision-focused learning. 

We define the \textit{scale} of a vector to refer to any scalar factor (including but not limited to its norm) by which the vector may be multiplied. This abstraction allows us to analyze the directional components independently of scale. We now present another essential property of optimization mappings: their scale invariance. 
\begin{property} \label{prop:scaling}
    Let us consider an optimization mapping $f:\Theta \mapsto \mathcal{C}$. For all cost vectors $\vtheta \in \Theta$ and positive scalar $\alpha > 0$, one has $f(\alpha \cdot \vtheta) = f(\vtheta)$.
\end{property}

Consider  a cost vector $\vtheta \in \Theta$ of stability radius $\eta$. Then according to property \ref{prop:scaling}, for any perturbation $\bm{\delta} \in \Theta$ and a positive scalar $\alpha$, one has: $f(\alpha \cdot \vtheta + \bm{\delta})=f( \vtheta + \alpha^{-1} \cdot \bm{\delta})$. Hence if $\alpha^{-1} \cdot \bm{\delta}$ is within distance $\eta$ of $\vtheta$, then one has $f(\alpha \cdot \vtheta + \bm{\delta})=f(\vtheta)$. We conclude that the stability radius of $\alpha \cdot \vtheta$ is $\alpha \cdot \eta$. Hence, a cost vector has its stability radius proportional to its scale and the following property holds.

\begin{property} \label{prop:scale_comp}
    Consider two cost vectors $\vtheta, \ \bm{\delta} \in \Theta$. The decision $f(\vtheta + \bm{\delta})$ depends on a comparison of scale between $\vtheta$ and $\bm{\delta}$. If $\vtheta$ is of arbitrarily greater scale than $\bm{\delta}$ then $f(\vtheta+\bm{\delta})=f(\vtheta)$. Conversely if $\vtheta$ is of arbitrarily lower scale than $\bm{\delta}$ then $f(\vtheta+\bm{\delta})=f(\bm{\delta})$. 
\end{property}

Most decision-focused learning techniques differentiate the optimization mapping through additive cost perturbations. We examine in the following section how solution stability, reflecting the mapping’s behavior under such perturbations, can negatively impact the learning process.

\section{Detrimental Effects of Solution Stability on the Decision-Focused Learning Process}

We study in this section the consequences of solution stability on the decision-focused learning process.

Perturbation-based DFL techniques rely on a perturbed decision mapping $\Tilde{f}: \Theta \mapsto \mathcal{C}$ that introduces an additive perturbation to the input cost vector before evaluation of the corresponding optimization mapping. In other words, for a cost vector $\vtheta \in \Theta$ and a cost perturbation $\bm{\delta} \in \Theta$, the corresponding perturbed decision is:
\begin{equation} \label{eq:perturbed_decision_mapping}
    \Tilde{f}(\vtheta) = f(\vtheta + \bm{\delta})
\end{equation}
DFL techniques leverage the perturbed optimization mapping to identify relevant descent directions minimizing the loss function by exploration in the neighborhood of $f(\vtheta)$ and comparison of perturbed decisions $\Tilde{f}(\vtheta)$ with $f(\vtheta)$. Hence, it is necessary that all perturbed decisions remain close to but different from $f(\vtheta)$. 

According to property \ref{prop:scale_comp}, the perturbed decision $\Tilde{f}(\vtheta)$ depends on a comparison of scale between $\vtheta$ and the perturbation $\bm{\delta}$:
\begin{itemize}
    \item If $\vtheta$ is of arbitrary greater scale than $\bm{\delta}$, the perturbed decision is equal to the exact decision $f(\vtheta)$. In this situation, the neighborhood around $f(\vtheta)$ is not effectively explored, making it hardly possible to identify a relevant descent direction.
    \item If $\vtheta$ is of arbitrary lower scale than $\bm{\delta}$, the perturbed decision is equal to $f(\bm{\delta})$. Because $f(\bm{\delta})$ generally lies far from the neighborhood of $f(\vtheta)$, the resulting descent direction may be misleading. 
\end{itemize}

We conclude that the perturbation scale must be chosen proportional to the scale of the cost vector estimate for the decision-focused learning process to be effective. In practice, the scale of perturbation depends on the chosen parameters value of each DFL techniques during the learning phase. It is therefore obvious that setting hyperparameters on an appropriate scale is necessary to ensure that gradient estimates remain meaningful throughout the training phase. 

However, this alone does not prevent the adverse effects of perturbation mismanagement during learning. Indeed, hyperparameters value remains constant throughout the training phase, while the weights of the ML model and its outputs evolve over the process, potentially altering their scale. On one hand, the perturbation scale remains fixed once hyperparameter values are chosen; on the other hand, the scale of estimated cost vectors may increase or decrease throughout training. 

Consequently, shifts in the relative scale between perturbations and cost vectors during the learning phase can lead to detrimental effects. In the following, we characterize these effects and analyze how mismanaging solution stability impacts each state-of-the-art perturbation-based DFL techniques. 

\section{Impacts of Solution Stability on the Learning Process}

We propose in this section an in-depth study of state-of-the-art DFL techniques with regard to the consequences of solution stability on their respective differentiation approach. A distinction is first established between two learning paradigms: imitation and experience. We then show that state-of-the-art DFL techniques that are expected to learn by experience may see their performances degraded by the effect of solution stability, leading either to non-differentiability or to a paradigm shift toward imitation. 

\subsection{Learning from Decisions by Imitation or Experience}

\citet{Bengio_2021} classify DFL techniques based on whether they learn by imitation or by experience. In the imitation setting, optimal decisions are inferred by replicating expert-provided behavior, without direct optimization of a performance measure. In contrast, experience-based learning involves iterative trial-and-error, guided by rewards and penalties, with the explicit goal of optimizing a defined performance metric. Our objective is to demonstrate that, without controlling solution stability, DFL techniques designed for learning through experience become either inefficient or collapse into imitation-based learning. We regard this outcome as equally undesirable, since broadly applicable methods, intended to work with any loss function, would be reduced to merely replicating existing solutions regardless of the performance they are meant to enhance.

We first identify a characteristic sufficient to classify a DFL technique as learning by imitation. To this end, we examine the Fenchel–Young (FY) loss, which is widely regarded as a standard imitation loss function \citep{Dalle_2022}. The FY loss is introduced by \citet{Blondel_2020} and corresponds to the cost comparison of the ground-truth decision $f(\bar{\vtheta})$ and a perturbed estimate decision $\Tilde{f}(\vtheta)$. It can be differentiated with regards to $\vtheta$ such that:
\begin{equation}
    \nabla_{\vtheta} L^{FY}(\vtheta,\bar{\vtheta})= f(\bar{\vtheta}) - \Tilde{f}(\vtheta) 
\end{equation}
Gradient $\nabla_{\vtheta} L^{FY}(\vtheta,\bar{\vtheta})$ indicates a descent direction toward a local minimum. Following this direction produces cost vector estimates whose optimization mapping best replicates the ground-truth decisions. Hence, the FY loss is clearly associated with a learning process by imitation.

In the following we show that under certain conditions, the gradient estimates of several state-of-the-art DFL techniques can resemble that of the Fenchel–Young loss, effectively categorizing them as learning processes by imitation.

\subsection{Smart Predict-then-Optimize}

Let us examine the Smart-Predict-then-Optimize (SPO) loss introduced by \citet{Elmachtoub_2022}. The SPO loss is presented as a differentiable surrogate for the original regret loss defined in \eqref{eq:regret_loss}. Hence it should be expected that it corresponds to a straightforward learning process by experience dependent on the regret performance measure. This surrogate loss $L^{SPO}_{\alpha}$ is defined according to a positive scalar $\alpha$, and \citet{Elmachtoub_2022} determine that it admits a sub-gradient with regards to the $\vtheta$ such that:
\begin{equation}
    \nabla_{\vtheta} L^{SPO}_{\alpha}(\vtheta,\bar{\vtheta}) \approx f(\bar{\vtheta}) - f(\alpha \vtheta-\bar{\vtheta})
\end{equation}
For an arbitrary lower scale of $\alpha \vtheta$ compared to that of $\bar{\vtheta}$, one has $f(\alpha \vtheta-\bar{\vtheta}) = f(-\bar{\vtheta})$ according to Property \ref{prop:scale_comp}. Hence the gradient estimate of the SPO loss takes the form of the following difference in decisions: 
\begin{equation}
\label{eq:SPO_no_info}
\nabla_{\vtheta} L^{SPO}_{\alpha}  \approx   f(\bar{\vtheta}) - f(-\bar{\vtheta}) 
\end{equation}
In this case, all information related to the cost vector estimate $\vtheta$ is lost in the computation of the gradient estimate. In other words, this gradient estimate cannot indicate a descent direction starting from $\vtheta$ that would improve the training loss. Hence, differentiation is not efficient.

Conversely if the scale of $\alpha \vtheta$ is arbitrarily greater than that of $\bar{\vtheta}$, then $f(\alpha \vtheta-\bar{\vtheta}) =  f(\vtheta)$ according to Property \ref{prop:scale_comp}. Hence the gradient estimate of the SPO loss takes the form of the following difference in decisions: 
\begin{equation}
\label{eq:spo_imitation}
\nabla_{\vtheta} L^{SPO}_{\alpha}  \approx  f(\bar{\vtheta}) - f(\vtheta) \approx \nabla_{\vtheta} L^{FY}(\vtheta,\bar{\vtheta})
\end{equation}
In this scenario, the SPO loss gradient aligns with the Fenchel–Young loss gradient, causing the learning process to degenerate from experience-based learning to imitation-based learning.

We conclude that there could be a shift in learning behavior when training a model with the SPO loss function, depending on the stability of solution $f(\alpha \vtheta-\bar{\vtheta})$ and on the respective scales of $\alpha \vtheta$ and $\bar{\vtheta}$.

\subsection{Implicit Differentiation by Perturbations}

In this section, we investigate a category of DFL techniques that construct smooth approximations of the optimization mappings by adopting a probabilistic point of view. This approach covers three DFL techniques : Perturb-and-MAP introduced by \citet{Papandreou_2011}, Implicit Maximum Likelihood Estimator introduced by \citet{Niepert_2021} and Differentiable BlackBox introduced by \citet{Pogancic_2020}. These methods are considered to be "blackbox" since they depend neither on the optimization problem nor on the considered loss function. Hence we assume they are part of an ML model that learn from decisions by experience.

These methods map the cost vector $\vtheta$ to a probability distribution $p(\vy|\vtheta)$ over the solution set $\mathcal{C}$. For a given cost vector $\vtheta$, an approximation of the exact optimal decision $f(\vtheta)$ can be obtained from the expectation of decisions according to the probability distribution $p(\vy|\vtheta)$, that is denoted $\mathbb{E}_{p}[\vy|\vtheta] \approx f(\vtheta)$. The perturbed decision $\mathbb{E}_{p}[\vy|\vtheta]$ can be seen as the expected value vector of a distribution of decisions centered at $f(\vtheta)$. 

%Note that all previously established stability properties hold for the perturbed mapping $\mathbb{E}_{p}[y|\cdot]:\Theta \mapsto \mathcal{C}$.

Implicit differentiation by perturbation is originally introduced by \citet{Domke_2010}. Its application in decision-focused learning leads to the following results. Given a positive scalar $\beta > 0$, the training loss function can see its gradient with regards to $\vtheta$ be approximated such that:
\begin{equation}
    \nabla_{\vtheta}  L(\vx,\vz;\vw)  \approx  \mathbb{E}_{p}[\vy|\vtheta+\beta \cdot \nabla_{\vy} L(\vx,\vz;\vw)] - \mathbb{E}_{p}[\vy|\vtheta]
\end{equation}

All three previously mentioned approaches rely on this approximation for decision mapping differentiation. For all three methods, the distribution $p(\vy|\vtheta)$ is obtained by introduction of an additive perturbation to the cost vector. In practice, a finite number of sample are drawn from a random multivariate variable $\rvd$ of values in $\Theta$ to approximate the expectation of decisions such that $\mathbb{E}_{p}[\vy|\vtheta] = \mathbb{E}_{\rvd}[f(\vtheta+\rvd)]$. If the scale of all samples is arbitrarily greater than that of $\vtheta$, then one has $\mathbb{E}_{p}[\vy|\vtheta]$ being approximated by $\mathbb{E}_{\rvd}[f(\rvd)]$. In other words, if the perturbation is of greater scale than cost vector $\vtheta$, then we no longer study decision in the neighborhood of $\vtheta$ and all information relating to $\vtheta$ is lost. We conclude that this differentiation approach is no longer efficient if the scale of perturbation is higher than that of the cost vector estimate. 

We now study the difference of scale between cost vectors $\vtheta$ and $\beta \cdot \nabla_{\vy} L(\vx,\vz;\vw)$. If the scale of $\vtheta$ is arbitrarily greater, then according to Property \ref{prop:scale_comp} one has :
\begin{equation}
\label{eq:implicit_null_grad}
    \nabla_{\vtheta}  L(\vx,\vz;\vw)  \approx  \mathbb{E}_{p}[\vy|\vtheta] - \mathbb{E}_{p}[\vy|\vtheta] = 0
\end{equation}
Hence differentiation is hardly feasible since the neighborhood around $f(\vtheta)$ remains unexplored and no descent direction can be found. On the contrary, if the scale of $\vtheta$ is arbitrarily lower than that of $\beta \cdot \nabla_{\vy} L(\vx,\vz;\vw)$, then according to Property \ref{prop:scale_comp} one has:
\begin{equation}
\label{eq:implicit_imitation}
    \nabla_{\vtheta}  L(\vx,\vz;\vw)  \approx  \mathbb{E}_{p}[\vy|\nabla_{\vy} L(\vx,\vz;\vw)] - \mathbb{E}_{p}[\vy|\vtheta]
\end{equation}
Similarly to the Fenchel-Young loss, following this gradient estimate produces cost vector estimates whose optimization mapping best replicates the ground-truth decisions. Hence in this case, this approach to differentiation would be similar to that of a learning process by imitation.

For example, when studying the regret loss $L^r$ defined in \eqref{eq:regret_loss}, according to \eqref{eq:regret_diff_y} one would have:
\begin{equation}
    \nabla_{\vtheta} L^r(\vx,\bar{\vtheta};\vw)  \approx \mathbb{E}_{p}[\vy|\bar{\vtheta}] - \mathbb{E}_{p}[\vy|\vtheta] \approx \nabla_{\vtheta} L^{FY}(\vtheta,\bar{\vtheta})
\end{equation}
In this case, the regret loss gradient corresponds to the FY loss gradient. Hence we observe that a scale of $\vtheta$ arbitrarily lower than that of $\beta \cdot \bar{\vtheta}$ causes the learning process to degenerate from experience-based learning to imitation-based learning.

To conclude, it is essential for all introduced perturbations to be at scale with the cost vector estimate so that DFL techniques relying on implicit differentiation by perturbation remain efficient.

\subsection{Differentiable Perturbed Optimizers}

In this section, we investigate Differentiable Perturbed Optimizers \citep{Berthet_2020}. This technique is similar to implicit differentiation by perturbations since it is a blackbox approach that relies on the perturbation of the optimal decision estimate in order to explore its neighborhood and identify direction descent of improvement. Therefore, we assume it is part of a ML model that learns from decisions by experience.

The DPO approach considers a probability distribution on the perturbed cost vector $\vtheta$ rather than on the optimal decision $f(\vtheta)$. A perturbed optimizer $f_{\epsilon}(\cdot)$ is introduced as a deterministic function of the cost vector $\vtheta$ and a multivariate random variable $\rvd$. This random variable acts as an additive perturbation on the cost vector with a temperature $\epsilon$, such that $f_{\epsilon}(\vtheta)= \mathbb{E}[f(\vtheta+\epsilon \cdot \rvd)]$. Assuming that the random variable $\rvd$ has a density proportional to $\exp(-\nu(\rvd))$ with function $\nu$ being twice-differentiable, the perturbed optimizer $f_{\epsilon}(\cdot)$ has its derivative taking the form:
\begin{equation}
\label{eq:DPO_diff}
    \nabla_{\vtheta}  f_{\epsilon}(\vtheta) = \mathbb{E}[f(\vtheta+\epsilon \cdot \rvd)\cdot \nabla_{\rvd} \nu(\rvd)^\intercal] 
\end{equation}
This gradient is in turn integrated into the gradient estimation of the training loss according to \eqref{eq:L_gradient}. 

In practice, the random variable $\rvd$ follows either a Gaussian or Gumbel distribution law for which one has $\mathbb{E} [\nabla_{\rvd} \nu(\rvd)]=0$. In practice, gradient $\nabla_{\vtheta}  f_{\epsilon}(\vtheta)$ is estimated by means of a Monte-Carlo sampling over the random variable $\rvd$. If for all samples, cost vector $\vtheta$ is of arbitrarily greater scale than $\epsilon \cdot \rvd$ then according to Property \ref{prop:scale_comp} one has $f(\vtheta+\epsilon \cdot \rvd)$ being approximated by $f(\vtheta)$ and the gradient estimate is of zero value. In this case, the perturbation is too low to identify suitable descent decisions in the neighborhood of $f(\vtheta)$, hence differentiation is hardly feasible.

On the contrary, if $\vtheta$ is of arbitrarily lower scale than all samples then differentiation is possible, but some information is lost. Indeed, according to Property \ref{prop:scale_comp}, it follows that one has $f(\vtheta+\epsilon \cdot \rvd)$ being approximated by $f(\rvd)$ and:
\begin{equation}
\label{eq:DPO_no_info}
    \nabla_{\vtheta}  f_{\epsilon}(\vtheta)  = \mathbb{E}[f(\rvd)\cdot \nabla_{\rvd} \nu(\rvd)^\intercal]
\end{equation}
In other words, the perturbation is too great and the gradient estimate for the DPO approach loses all information on the cost vector estimate $\vtheta$. Hence it is not able to indicate a descent direction starting from $\vtheta$ that would improve the performances evaluated by the training loss.

We conclude that, for effective differentiation, the perturbation introduced in the DPO approach must be on the same scale as the cost vector estimate.

\section{Managing Solution Stability with Cost Regularization}
\label{sec:reg}

We have shown that the general learning behavior emerging from perturbation-based DFL techniques is deeply affected by the balance in scale between cost vectors and perturbations. If this balance is lost, learning may collapse or turn into imitation, regardless of the initial perturbation intensity set by hyperparameters. In this section, we explore ways to address this issue through cost regularization.

\subsection{Ensuring Cost Vectors Remains at Scale with Perturbations}

Managing solution stability to improve the reliability of the learning process while maintaining performances appears to be a significant challenge. According to property \ref{prop:perturbation_stability}, it is necessary that the stability radius of the cost vector estimate be on the same scale as that of the perturbation to ensure this outcome. To the best of our knowledge, no further information is available regarding the exact value that the stability radius should assume to maximize the efficiency of the differentiation process. 

We therefore propose imposing bounds on the stability radius of all estimates in order to control its scale without assigning an exact value. According to property \ref{prop:scaling}, the stability radius of a cost vector is directly proportion to its norm since it is a natural measure of its scale. Hence imposing bounds on the vector norm should constrain the stability radius. It is indeed the case for upper-bounds, as demonstrated by the following property:
\begin{property} \label{prop:bounding}
    Given a non-constant optimization mapping \( f: \Theta \to \mathcal{C} \), the stability radius of each cost vector in a bounded subset \( \sT \subset \Theta \) is uniformly upper-bounded.
\end{property}
\begin{proof}
First, we denote by $\mathcal{B}(\vtheta,\rho)$ the $\normltwo$ ball of radius $\rho$ in $\mathbb{R}^n$ centered on $\vtheta$. We consider a non-constant optimization mapping $f:\Theta \mapsto \mathcal{C}$. Let $\sT \subset \Theta$ be a bounded set, with $\rho$ an upper-bound on the norm of all cost vectors in that set. If there does not exist an upper-bound on the stability radius of each cost vector in that set, then there exists a cost vector $\vtheta^0 \in \sT$ such that its stability radius is greater than $2\rho$. Hence for all perturbation $\bm{\delta} \in \Theta$ with a norm lower than $2\rho$, one has $f(\vtheta^0 + \bm{\delta}) = f(\vtheta^0)$. 

Note that the set $\mathcal{B}(\bm{0},\rho)$ is a subset of $\mathcal{B}(\vtheta^0,2 \rho)$ since $\vtheta^0 \in \mathcal{B}(\bm{0},\rho)$. We therefore conclude that one has $f(\vtheta)=f(\vtheta^0)$, for all $\vtheta \in \Theta \cap \mathcal{B}(\bm{0},\rho) $. Hence according to property \ref{prop:scaling}, the optimization mapping $f$ is constant over the set $\Theta$. This is absurd since the optimization mapping $f$ is supposed to be non-constant, hence we conclude that the stability radius is uniformly upper-bounded over the bounded set $\sT \subset \Theta$. 
\end{proof}

However, there does not exist a similar general property on lower bounds with regards to the relation between vector norm and stability radius, as shown by the counter-example provided in section \ref{section:solution_stability}. Instead and as illustrated in section \ref{section:solution_stability}, it appears that the subset of cost vectors with lower-bounded stability radius is dependent on the optimization problem structure. 

We restrict our work to general regularization approaches dedicated to the correct management of solution stability in decision-focused learning, hence we focus in what follows on regularization to impose upper bounds on vectors norm and stability radius. Nonetheless, the development of regularization techniques to enforce lower bounds on the stability radius remains an open question for future research.

\subsection{Cost Regularization Approaches}

We propose to implement a regularization function applied to the exact cost vector estimate $\vtheta$ before cost perturbation. 

Consider a perturbed optimization mapping $\Tilde{f}: \Theta \mapsto \mathcal{C}$ as defined in \eqref{eq:perturbed_decision_mapping}. The regularization function is defined as a mapping $r: \Theta \to \Theta$ satisfying $f(r(\boldsymbol{\theta})) = f(\boldsymbol{\theta})$, so that the exact decision mapping remains unchanged. The role of $r$ is to rescale $\boldsymbol{\theta}$ to match the scale of the perturbation $\boldsymbol{\delta}$ introduced by the perturbed optimization mapping $\tilde{f}$, such that $\tilde{f}(r(\boldsymbol{\theta})) = f(r(\boldsymbol{\theta}) + \boldsymbol{\delta})$. 

The regularized decision-focused learning model is then described by:
\begin{equation}
\label{eq:reg_ML_model}
\vtheta = h_{\vv}(\vx), \quad \vy=\Tilde{f}\big(r(\vtheta)\big)
\end{equation}

\begin{definition}
    Let $r^n:\Theta \mapsto \Theta$ defines a $\normltwo$ normalization of the cost vector estimate, such that:
\begin{equation}
    r^n(\theta)=\frac{\vtheta}{||\vtheta||}, \quad \forall \theta \in \Theta
\end{equation}
\end{definition}
Regularization $r^n$ projects the cost vector estimates unto the unit sphere, thereby upper-bounding the value of their stability radius. According to the scale invariance property  \ref{prop:scaling}, one has $f(r^n(\vtheta))=f(\vtheta)$. 

In this first approach, for any cost vector $\theta \in \Theta$, the stability radius of $\alpha \cdot \theta$ is constant for all positive scalar $\alpha$. Hence, although this regularization does not assign a single value to the stability radius of all cost vectors, it does so for vectors sharing the same direction component. This partly contradicts our assumption that the stability radius should remain flexible in its precise value. 

We propose to consider another regularization in which the cost vector norm is upper-bounded rather than having its value set to $1$, thereby bounding the stability radius for a given direction component rather than assigning it an arbitrary value. 
\begin{definition}
    Let $r^p:\Theta \mapsto \Theta$ defines a regularization function smoothly rescaling cost vector estimates so that their $\normltwo$ norm does not exceed a given positive value $\kappa$: 
\begin{equation}
    r^p(\vtheta) = \frac{1}{1+\kappa^{-1} \left \| \vtheta \right \|} \cdot \vtheta
\end{equation}
\end{definition}
This regularization corresponds to a projection into an $\normltwo$ ball of radius $\kappa$ ensuring the intended balance between rescaling and flexibility outlined above. Moreover by scale invariance, one has $f(r^p(\vtheta))=f(\vtheta)$. 

This regularization preserves the ordering of vector norms, leaving cost vectors with norms well below $\kappa$ essentially unchanged while down-scaling those with larger norms. The choice of value for parameter $\kappa$ balances flexibility and regularization: larger values yield looser bounds on both the norm and the stability radius.

\section{Numerical Experiments}
\label{sec:num_exp}

We conduct experiments to validate the effectiveness of the proposed regularization approach. Designing an experimental setup that reliably captures training instability caused by perturbations in DFL methods is challenging, as such instability is neither systematic nor easily observable. That is why we propose two complementary sets of experiments: the first highlights general relevance, while the second provides direct evidence of the problem our approach resolves.

To begin, we present experiments on a well-established benchmark to illustrate the broad applicability and potential benefits of our approach, even though these results do not conclusively demonstrate that improved performance stems from enhanced solution stability. To address this gap, we then propose an in-depth study with a dedicated experiment deliberately reproducing solution stability issues, making it evident that our method specifically targets and mitigates this problem. The code is directly derived from the original work of \citet{Mandi_2024} and is available for reproducibility purposes \citep{repo}.

\subsection{Experimentation setup and statistical testing}
\label{section:exp}

All experiments and results can be reproduced with the code provided in the indicated Git repository \citep{repo}.

For each instance of each problem, we consider a learning model described by \eqref{eq:ML_model} such that $h_{\vv}(\cdot)$ is a fully connected neural network. For each experiment, ten trials are conducted during both validation and testing phases, with network weights initialized using distinct seeds ranging from 0 to 9. Performance is assessed based on the average regret over the test dataset across these ten training runs. The number of training epochs is fixed at 30, and a grid search is performed to identify the optimal hyper-parameter set for each DFL technique using a designated evaluation dataset. 

We propose in table \ref{tab:grid_search} the value range of each parameter involved in one of the studied learning approaches in section \ref{sec:num_exp}.

\begin{table*}[h]
    \centering
    \begin{tabular}{c|c|c}\toprule
        Hyper-parameter & DFL technique & Range \\ \midrule
        learning rate & All & $\{10^{-4}, 10^{-3}, 10^{-2}, 10^{-1}, 1 \} $ \\
        $\kappa$ & All & $\{1,10^2,10^4\}$ \\
        $\alpha$ & SPO & $\{0.5, 2, 10\} $ \\
        $\beta$ & DBB & $\{0.1, 1, 10\} $ \\
        $\epsilon$ & DPO & $\{0.1,0.5,1\}$ 
    \end{tabular}
    \caption{Value table for the hyper-parameter tuning of all learning approaches}
    \label{tab:grid_search}
\end{table*}

All parameters range have been selected according to pre-tests leading to the dismissal of exceedingly large or small values, based on numerical experiments proposed by \citeauthor{Mandi_2024} that designed this benchmark. Regarding the range of values for the parameter $\kappa$ in regularization $r^p$, they act as a measure of the intensity of the regularization, with each value corresponding respectively to a strict, moderate or low regularization. The number of possible values for each parameter range is assumed to be fair given the extensive computation time required to train each decision-focused learning model. 

For each optimization problem and its associated datasets, the performances of all DFL techniques are reported using their respective optimal hyperparameter configurations, evaluated according to a regret-based metric. For each individual model, the reported performance corresponds to the average regret obtained over ten training runs with random seeds ranging from 0 to 9. Results are summarized in tables \ref{tab:perf} and \ref{tab:exp_te}. 

We provide information relative to the statistical relevance of our claim stating that the regularized DFL techniques perform better than their standard counterpart. 

To assess whether the proposed regularization consistently outperforms the baseline under identical training conditions, we conduct a paired non-parametric analysis across matched random seeds, using the best hyperparameter configurations identified for each model and evaluating performance via the regret computed on the test dataset. We apply a Wilcoxon signed-rank test with a one-sided alternative hypothesis and excluding ties to determine whether the best-performing method consistently surpasses its counterpart, for each problem, dataset and DFL technique.

\subsection{Benchmark Study}
\label{sec:benchmark_study}

We propose a first study based on the state-of-the-art review by \citet{Mandi_2024}. The experiments consist in two combinatorial optimization problems with publicly available datasets, each with three instances: the synthetic shortest path problem (SP) and the set matching problem (SM). We aim to evaluate the relevance of regularization for the smart predict-then-optimize loss, the implicit differentiation by perturbation and the differentiable perturbed optimizers. 

To this end, the performance of the SPO, DBB, and DPO approaches is evaluated in terms of regret expressed as a percentage, both with and without regularization. The considered loss function for the DPO and DBB approach is the regret loss $L^r$ defined in \eqref{eq:regret_loss}. For context and comparison, we also report the results from predictions obtained with a standard mean-squared error (MSE) loss. 

We replicate the experimental setup proposed by \citet{Mandi_2024}. Models are trained over multiple initialization seed. Their hyperparameters are optimized according to an evaluation dataset by grid search, and their performances are established on a test dataset.  

\begin{table*}[h]
    \centering
    
    \caption{ Regret as percentage (\%) for various problems and DFL techniques.}
    
    \setlength{\tabcolsep}{3pt}
    \begin{tabular}{c|c|ccc|ccc|ccc}
        & MSE & SPO & SPO-$r^n$ & SPO-$r^p$ & DPO & DPO-$r^n$ & DPO-$r^p$ & DBB & DBB-$r^n$ & DBB-$r^p$ \\ \midrule
        SP1 & 15.29 & 21.24 & \textbf{15.40} & \textbf{15.40} & 17.73 & \textbf{16.08} & 16.30 & 16.55 & \textit{17.92} &\textbf{16.41} \\ 
        SP2 & 10.12 & \textbf{10.07} & \textit{10.12} & \textit{10.40} & 11.83 & \textbf{10.55} & 11.18 & 11.92 & 11.32 &\textbf{10.83} \\ 
        SP3 & 19.79 & 8.43 & \textbf{7.67} & \textit{8.50} & 125.57 & \textbf{13.10} & 15.72 & 10.15 & \textbf{9.63} & \textit{11.29}  \\ \midrule
        SM1 & 92.09 & 91.72 & 90.38 & \textbf{90.09} & 92.08 & \textit{92.27} &\textbf{92.03} & 91.53 & 91.16 &\textbf{90.62} \\ 
        SM2 & 92.23 & 91.90 & \textbf{89.31} & 89.66 & 92.35  & \textit{92.66} &\textbf{92.33} & 91.19 & 90.90 &\textbf{90.17} \\ 
        SM3 & 91.87 & \textbf{88.02} & \textit{88.79} & \textit{88.14} & 92.35 & 92.33 &\textbf{92.05} & 90.19 & \textit{91.13} &\textbf{88.97} \\ 
    \end{tabular}
    \label{tab:perf}
\end{table*}

Table \ref{tab:perf} presents the best performance, in terms of regret, of each DFL technique under each regularization method. The best results for each DFL technique and optimization problem are shown in \textbf{bold}, while regularized DFL techniques performing worse than their unregularized counterparts are shown in \textit{italics}. The corresponding p-values are reported in tables \ref{tab:p_value}.

\begin{table*}
    \centering
    \caption{Statistical testing for the general study.}
    \begin{tabular}{c|c|c|c}
        & SPO& DPO & DBB  \\ \midrule
        SP1 & 0.05 & 0.001 & 0.001  \\ 
        SP2 &  \textit{0.001} & 0.001  & 0.001  \\ 
        SP3 &  \textit{0.07} & 0.001  & \textit{0.002} \\ \midrule
        SM1 & 0.03 & 0.15 & 0.05 \\ 
        SM2 &  0.05 & 0.32 & 0.03 \\ 
        SM3 &  \textit{0.46} & 0.53 & 0.09\\
    \end{tabular}
    \label{tab:p_value}
\end{table*}

Values in \textit{italics} indicate p-values testing whether the standard model performs better, while the remaining values test whether the regularized model is superior. The consistently low p-values observed across experiments support our claim that the regularized DFL techniques consistently outperform their standard counterparts in most cases.

Projection-based regularization $r^p$ appears more reliable than normalization-based regularization $r^n$, as it consistently outperforms all considered DFL techniques, with statistical supporting this claim as shown in table \ref{tab:p_value}. We conclude that introducing a regularization strategy that balances rescaling and flexibility is essential for consistently improving perturbation-based DFL techniques.

We cannot assume beforehand whether certain problems or datasets exhibit instability during DFL training. Our hypothesis is that, when instability is present, regularization significantly improves performance. Conversely, in cases where training remains stable, regularization offers minimal advantage. The experiments that follow confirm this hypothesis. 

\subsection{In-Depth Study}
\label{sec:te}

We present an in-depth study of solution stability issues, by introduction of a novel toy problem to illustrate the impact of poor solution stability management on decision-focused learning.

\begin{figure}[h]
    \centering
    \includegraphics[width=0.45\textwidth]{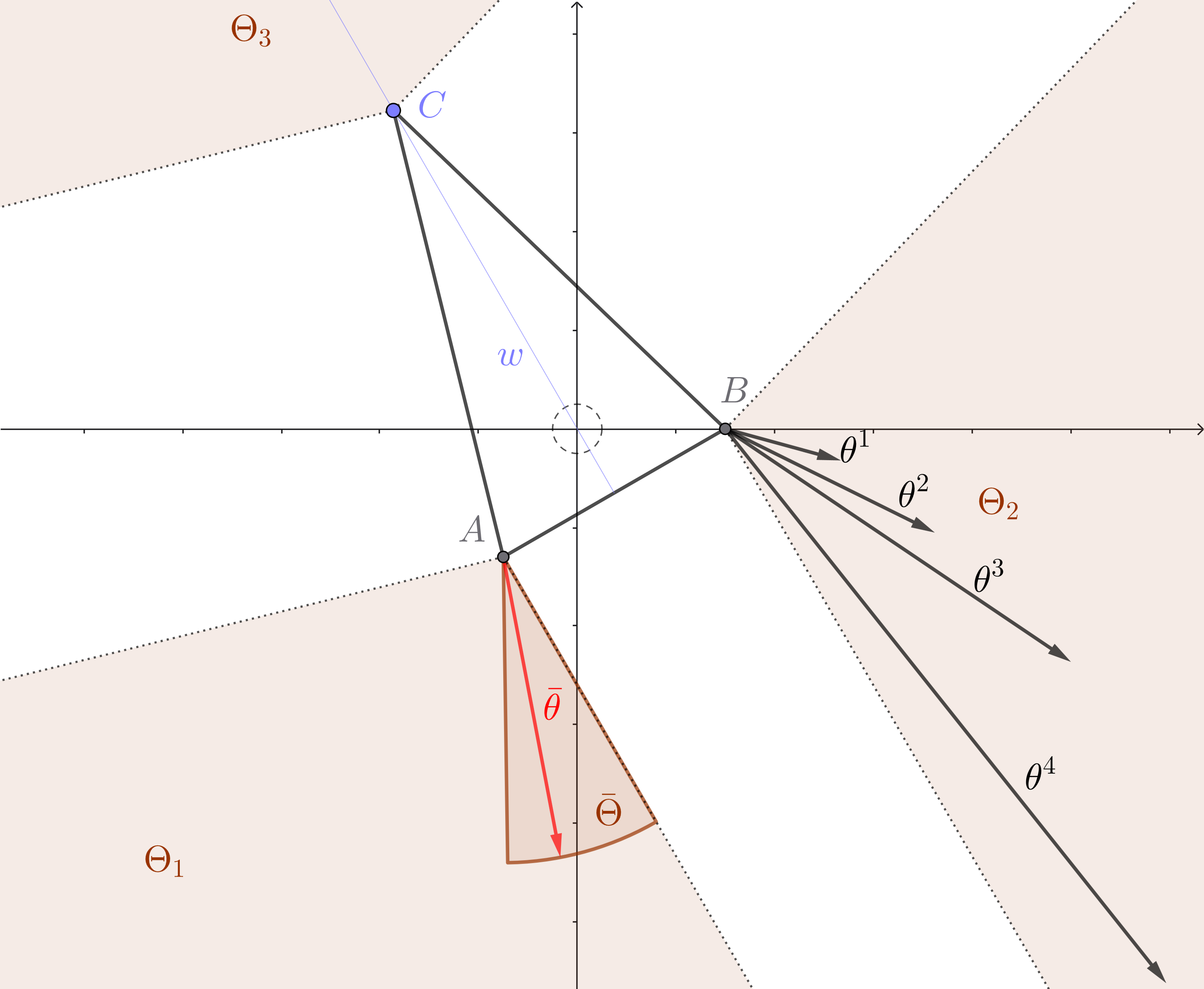} % first figure itself
    \caption{Toy Problem Illustration} \label{fig:TE}
\end{figure}

We consider a linear programming model illustrated by figure \ref{fig:TE}. In that problem, one has three vertex $A=(\sin(\frac{4 \pi}{3}),\sin(\frac{4 \pi}{3}))$, $B=(\cos(0),\sin(0))$ and $C$ along the half-axis $\{\alpha  \cdot (\sin(\frac{2 \pi}{3}),\sin(\frac{2 \pi}{3}))\in \mathbb{R}^2 | \ \ \alpha \geq 0 \}$. In that problem, one has $\Theta=\mathbb{R}^2$ and the linear objective is maximized for decisions in the convex hull $\mathcal{C}$ of vertex $A$, $B$ and $C$. The optimal decision is equal to $A$, $B$ or $C$, depending on the cone $\Theta_1$, $\Theta_2$ or $\Theta_3$ to which the directional component of the cost vector belongs.  

The corresponding optimization problem is reduced to:
\begin{align}
    \max \ \ & \vtheta^\intercal \cdot x \\
    \text{s.t.} \ & x \in \{A,B,C\}
\end{align} 
We examine two categories of problems, corresponding to three possible positions of vertex \(C\) along the \(w\) half-axis, with $\alpha$ of value $10$ or $100$. For each category, instances are generated by rotation of the polytope around the origin. 

For each problem category, a dataset of 300 instances is generated. It is then divided into a training dataset, an evaluation dataset and a test dataset of 100 instances each. Each instance $i$ is obtained by random generation of a true cost vector $\bar{\vtheta}^i$ in set $\{(\cos(\eta),\sin(\eta)) \in \mathbb{R}^2 | \ \eta \in [|\frac{3\pi}{2},\frac{7 \pi}{6}|]\}$ and of a random rotation angle $\zeta^i \in [|0,2\pi|]$. Then, for a rotation matrix $M^{\zeta}$ of angle $\zeta$ in two dimensions, the exact optimization problem for each instance is: 
\begin{align}
    \max \ \ & (M^{\zeta^i} \bar{\vtheta}^i)^\intercal \cdot x \\
    \text{s.t.} \ & x \in \{M^{\zeta^i}A, M^{\zeta^i}B, M^{\zeta^i}C\}
\end{align}
For each instance, the true cost vector \(\bar{\theta}\) is within set \(\bar{\Theta} \subset \Theta_1\), with a corresponding optimal decision \(A\). Likewise for each instance, decision $C$ is greatly detrimental with regards to cost, as it is diametrically  opposed to the true cost vector \(\bar{\theta}\) in set $\bar{\Theta}$. For each category, the ML model is trained on a dataset comprising all related instances where the input consists of rotation angles and the output consists of true cost vectors. 

This task is prone to degeneration under the DPO approach. For a cost perturbation \(\rvd\) drawn from a Gaussian distribution, the corresponding regret loss gradient is:
\begin{equation}
    \nabla_{\vtheta} L^r(\vx,\bar{\vtheta};\vv) = \bar{\vtheta}^\intercal \mathbb{E}\big[f(\vtheta+\epsilon \rvd)\cdot \rvd^\intercal\big].
\end{equation}
The gradient computation for this loss relies on a neighborhood search, since it captures how variations in the cost vector shift the solution from sub-optimal choices toward cost-maximizing ones. This approach becomes problematic in the present setting, because decisions A and B lie directly adjacent to the significantly worse decision C. The large cost discrepancy between these options is likely to introduce instability in the differentiation process, as observed in our numerical results.

We compare for each problem category the performances of an ML model trained with the DPO approach, with and without regularization. We restrict our study to the projection regularization $r^p$ since it appears it is the most efficient in our previous experiments. Our experimental setup is similar to that of the previous experiments, with details provided in section \ref{section:exp}. 

\begin{table}[h]
    \centering
    
    \caption{Regret (\%) for each category and method.}
    \setlength{\tabcolsep}{3pt}
    \begin{tabular}{c|cc}
        & Cat. 1 & Cat. 2  \\ \midrule
        DPO & 47.28 & 271.14  \\
        DPO-$r^p$  & \textbf{20.11} & \textbf{52.34} \\
    \end{tabular}
    \label{tab:exp_te}
\end{table}

Results are reported in Table~\ref{tab:exp_te} regarding the best performance in terms of regret for the DPO approach, both with and without regularization. The corresponding p-values are reported in tables \ref{tab:stat_te}.

\begin{table}[h]
    \centering
    \caption{Statistical testing for the in-depth study.}
    \begin{tabular}{c|ccc}
        & Cat. 1 & Cat. 2 &  \\ \midrule
        p-value & 0.001 & 0.001\\
    \end{tabular}
    \label{tab:stat_te}
\end{table}

Across both problem categories, the regularized formulation consistently achieves superior performance compared with the standard one, with statistical testing supporting that claim as shown in table \ref{tab:stat_te}. 

\begin{figure}[!htbp]
  \centering

  % Each subfigure is half the text width minus some gutter
  \newcommand{\imgw}{0.32\linewidth}

  \begin{subfigure}{\imgw}
    \centering\includegraphics[width=\linewidth]{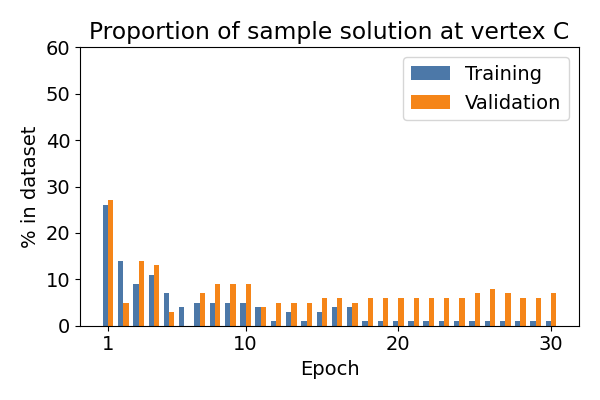}
    \caption{DPO model at seed 0}
  \end{subfigure}\hfill
  \begin{subfigure}{\imgw}
    \centering\includegraphics[width=\linewidth]{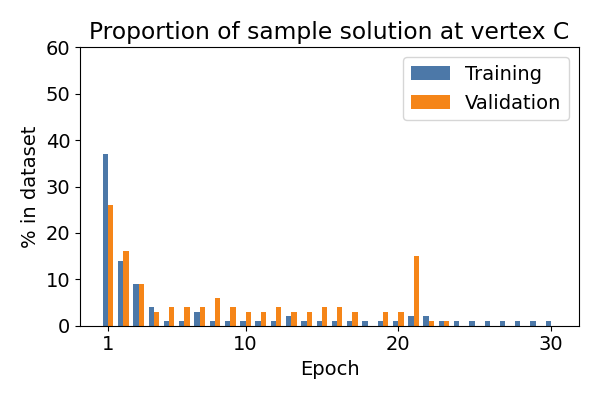}
    \caption{Regularized DPO model at seed 0}
  \end{subfigure}
  \begin{subfigure}{\imgw}
    \centering\includegraphics[width=\linewidth]{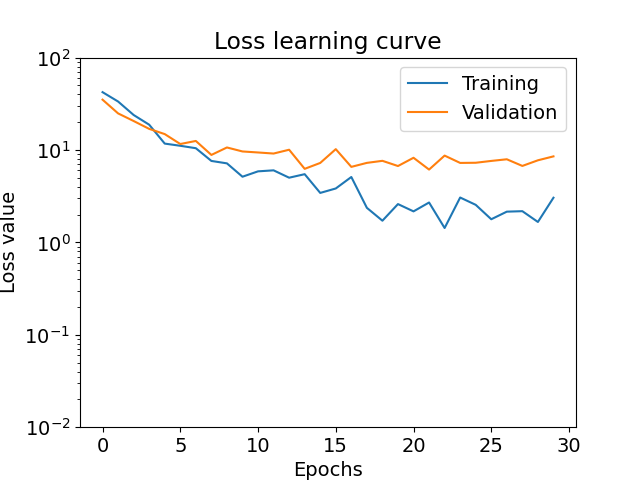}
    \caption{DPO model at seed 0}
  \end{subfigure}\hfill
  \par\vspace{0.5em}
  
  \caption{Problem category 2, seed 6: proportion of samples at vertex C and DPO model loss values.}
  \label{fig:DFL_te}
\end{figure}

This performance gap can be attributed to stability issues inherent to the unregularized model. Figures in section \label{section:te} support this claim: they display for each problem category and each random seed, the proportion of samples whose optimal decision lies at vertex~C throughout training, computed on both the training and evaluation datasets and shown for both the standard and regularized variants. We provide an illustration with figure \ref{fig:DFL_te}.

For the regularized DPO approach, this proportion in both the training and validation dataset decreases during learning and eventually reaches zero, as expected since assigning a decision at vertex~C entails a substantial increase in the loss.

In contrast, under the standard DPO approach, this proportion in the training dataset does not consistently converge to zero, which naturally leads to degraded performance compared to the regularized DPO model. Even in cases where it does vanish in the training set, the corresponding proportion in the validation dataset often remains strictly positive and plateaus at levels substantially higher than those observed during training. This behavior is indicative of overfitting: the model fails to reliably learn to map cost vectors outside $\Theta_3$ and instead overfits the few training samples exhibiting this configuration.

We further report the evolution of the DPO loss for the standard model to highlight that this overfitting is atypical. Despite the phenomenon described above, the evaluation loss does not display the usual increase associated with conventional overfitting. Rather than learning from the samples and subsequently overfitting them, the model is unable to learn from certain cases altogether and immediately overfits these specific instances.

Finally, this pathological behaviour becomes more pronounced in category~2, where decisions at vertex~C are even more detrimental, amplifying the model's instability. By contrast, no such behaviour is observed for the regularized model, which effectively mitigates these instabilities and thereby achieves superior performance.

We conclude that the unregularized model fails to reliably learn to map cost vectors outside $\Theta_3$ and instead can only overfits the training samples exhibiting this pattern.

Finally, this pathological behaviour becomes more pronounced in category~2, where decisions at vertex~C are even more detrimental, amplifying the model's instability. By contrast, no such behaviour is observed for the regularized model, which effectively mitigates these instabilities and thereby achieves superior performance.

\section{Conclusion}

In decision-focused learning, a key challenge arises from the solution stability of discrete optimization mappings when subjected to additive perturbations. Using theoretical insights from stability analysis in combinatorial optimization, our analysis revealed that shifts in the relative scale between perturbations and cost vectors in decision-focused learning  can negatively impact performance. To address this, we proposed regularizing the cost vector estimates during training. By striking a balance between rescaling and flexibility, our designed regularization approach consistently improves performances.

Our work paves the way for multiple directions of future research. Alternative perturbations such as multiplicative ones \citep{Dalle_2022} could be investigated since they may influence the differentiation process in other ways. Moreover, exploring how the problem’s structure influences solution stability could further help address solution stability issues.

\bibliographystyle{apalike}
\bibliography{bibfile}

\appendix

\section{Example to illustrate solution stability in decision-focused learning}
\label{section:solution_stability}

Consider the following problem in two dimensions, with linear costs parameterized by a vector $\vtheta \in \mathbb{R}^2$, and denoted $P(\vtheta)$:
\begin{align}
    P(\vtheta) : \min \ \ & \vtheta_1 \cdot \vx_1 + \vtheta_2 \cdot \vx_2 & \\
    & \vx_1 + \vx_2 = 1 & \\
    & 0 \leq \vx_i \leq 1, & \forall i \in \{1,2\}
\end{align}
In this simple problem, the optimal solution depends entirely on the order of costs. Indeed, the optimal solution is $\vx=(1,0)$ (resp. $\vx=(0,1)$) for all cost vectors such that $\vtheta_1 < \vtheta_2$ (resp. $\vtheta_1 > \vtheta_2$). Otherwise for all cost vectors such that $\vtheta_1 = \vtheta_2$, any solution $\vx \in \{(a,1-a) ;\; a \in [0,1] \} $ is optimal. 

Consider the set valued optimization mapping $F(\cdot)$ associated to problem $P$. The set $F(\vtheta)$ is a singleton $\{(1,0)\}$ (resp. $\{(0,1)\}$) for all cost vectors such that $\vtheta_1 < \vtheta_2$ (resp. $\vtheta_1 > \vtheta_2$), and is equal to $\{(a,1-a) ;\; a \in [0,1] \} $ for all cost vectors such that $\vtheta_1 = \vtheta_2$. Hence $F(\vtheta)$ is a singleton for all cost vectors $\theta \in \mathbb{R}^2$ except for those with a direction component along vector $(1,1)$. We conclude that $F(\cdot)$ is a singleton almost everywhere.

Consider now the optimization mapping $f : \mathbb{R}^2 \mapsto [0,1]^2$ associated to problem $P$, such that $f(\vtheta)$ is an optimal solution to problem $P(\vtheta)$ for a cost vector $\vtheta \in \mathbb{R}^2$. One can observe the invariance property for this particular example. For all cost vectors $\vtheta \in \mathbb{R}^2$ and positive scalars $\alpha >0$, one has $f(\alpha\cdot \vtheta)=f(\vtheta)$ since the order of costs remains the same for vectors $\vtheta$ and $\alpha\cdot \vtheta$. 

We now determine an explicit formulation to compute the stability radius of any cost vector for that particular problem. To that end, consider a cost vector $\vtheta \in \mathbb{R}^2$ such that $\vtheta_1 > \vtheta_2$ and a perturbed cost vector $\vtheta + \bm{\delta} \in \mathbb{R}^2$. The order of costs is identical for these two vectors if the $\normltwo$ norm of $\bm{\delta}$ is lower than $\vtheta_1 - \vtheta_2$, such that the first element of $\vtheta + \bm{\delta}$ cannot get smaller than the second element of that perturbed cost vector. Likewise, suppose that $\bm{\delta}=(0,\vtheta_1 - \vtheta_2 +\epsilon)$ for a given positive scalar $\epsilon >0$. This vector is of $\normltwo$ norm $\vtheta_1 - \vtheta_2+\epsilon$, and one has $\vtheta + \bm{\delta}= (\vtheta_1,\vtheta_1+\epsilon)$, hence the orders of cost of  $\vtheta$ and $\vtheta + \bm{\delta}$ are different. We deduce that $\vtheta$ and $\vtheta + \bm{\delta}$ are of different orders of costs if and only if $\bm{\delta}$ is of greater norm than $\vtheta_1 - \vtheta_2$. 

A symmetrical result can be found for a cost vector $\vtheta \in \mathbb{R}^2$ such that $\vtheta_1 < \vtheta_2$, such that $\vtheta$ and $\vtheta + \bm{\delta}$ are of different orders of costs if and only if $\bm{\delta}$ is of greater norm than $\vtheta_2 - \vtheta_1$. Since the order of costs for any vector $\vtheta$ determines its optimal solution $f(\vtheta)$ for problem $P$, we conclude that the stability radius of optimization mapping $f(\cdot)$ at $\vtheta$ is equal to $|\vtheta_1 - \vtheta_2|$. 

Consider now the sequence of cost vectors $\big( (t+\frac{1}{n}, t) \big)_{n>1}$ for a positive scalar $t > 0$. All cost vectors in that sequence have an $\normltwo$ norm greater than $t$. The associated stability radius sequence is $(\frac{1}{n})_{n>1}$, hence it converges to zero, independently of the scalar $t$. It follows in this particular case that, no matter the scale, there exist cost vectors in $\mathbb{R}^2$ with a stability radius of arbitrarily low value. This serves as a counter-example to the hypothesis, formulated in section \ref{sec:reg}, that a positive lower bound on the norm of cost vectors implies a positive lower bound on the associated stability radius. In other words, even if the norm of the cost vectors is bounded below, it does not guarantee the existence of a positive lower bound for the stability radius.

In this particular case, the subset of cost vectors in $\mathbb{R}^2$ with a lower bound $\underline{\kappa}$  on their stability radius is defined by $\{ \vtheta \in \mathbb{R}^2 ; \; |\vtheta_1 - \vtheta_2|\geq \underline{\kappa} \}$. We conclude that a subset of cost vectors with a lower bound on their stability radius can be identified by analysis of the optimization problem structure.

\section{Detailed numerical experiments: toy-example for regularized decision-focused learning}
\label{section:te}

We compare the standard and regularized DPO approach on two problem categories. Results, in terms of regret value, are reported in table \ref{tab:exp_te}: across both problem categories, the regularized formulation consistently achieves superior performance compared with the standard one, with statistical testing supporting that claim as shown in table \ref{tab:stat_te}. 

\begin{figure}[!htbp]
  \centering

  % Each subfigure is half the text width minus some gutter
  \newcommand{\imgw}{0.32\linewidth}

  \begin{subfigure}{\imgw}
    \centering\includegraphics[width=\linewidth]{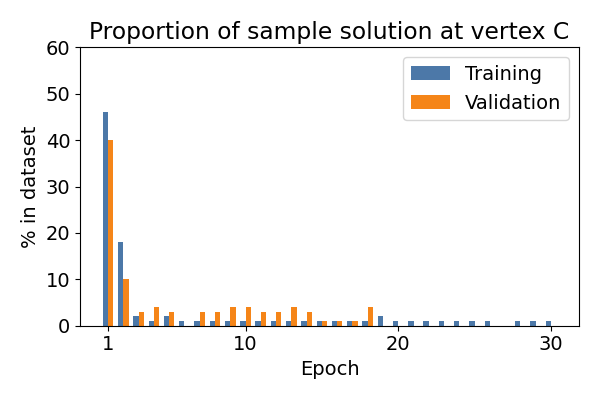}
    \caption{Regularized DPO model at seed 0}
  \end{subfigure}
  \begin{subfigure}{\imgw}
    \centering\includegraphics[width=\linewidth]{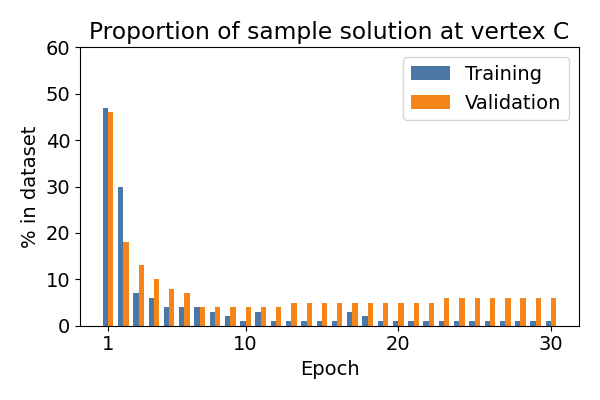}
    \caption{DPO model at seed 0}
  \end{subfigure}\hfill
  \begin{subfigure}{\imgw}
    \centering\includegraphics[width=\linewidth]{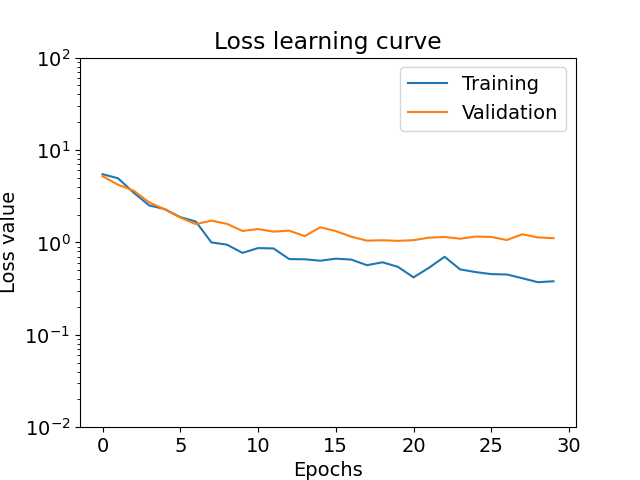}
    \caption{DPO model at seed 0}
  \end{subfigure}\hfill
  \par\vspace{0.5em}

  \begin{subfigure}{\imgw}
    \centering\includegraphics[width=\linewidth]{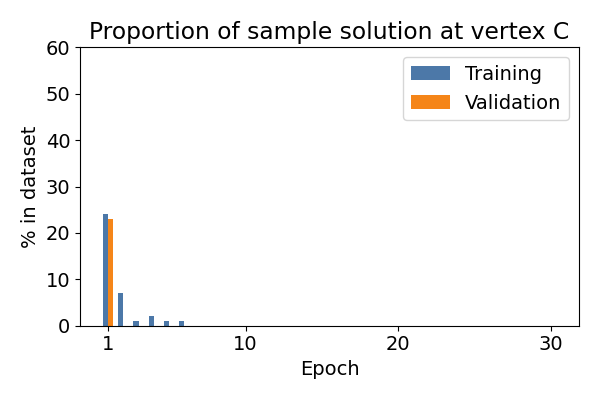}
        \caption{Regularized DPO model at seed 1}
  \end{subfigure}
  \begin{subfigure}{\imgw}
    \centering\includegraphics[width=\linewidth]{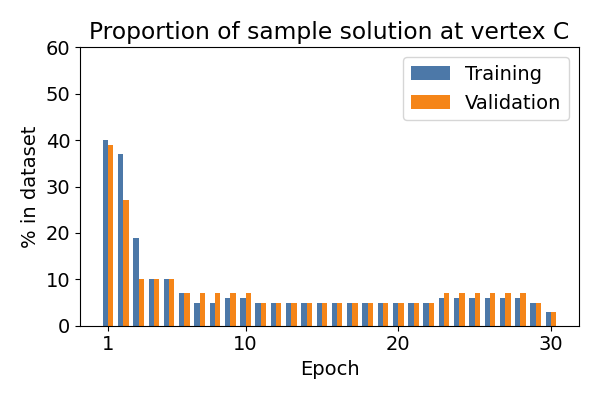}
    \caption{DPO model at seed 1}
  \end{subfigure}\hfill
  \begin{subfigure}{\imgw}
    \centering\includegraphics[width=\linewidth]{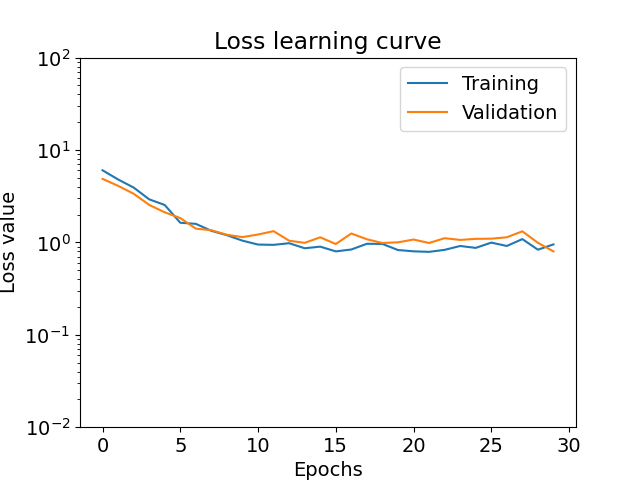}
    \caption{DPO model at seed 1}
  \end{subfigure}\hfill
  \par\vspace{0.5em}

  \begin{subfigure}{\imgw}
    \centering\includegraphics[width=\linewidth]{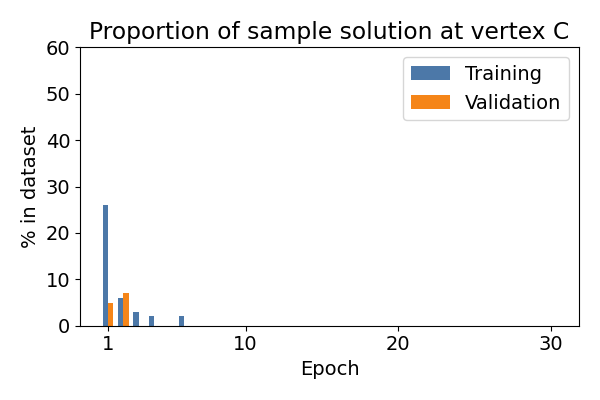}
        \caption{Regularized DPO model at seed 2}
  \end{subfigure}
  \begin{subfigure}{\imgw}
    \centering\includegraphics[width=\linewidth]{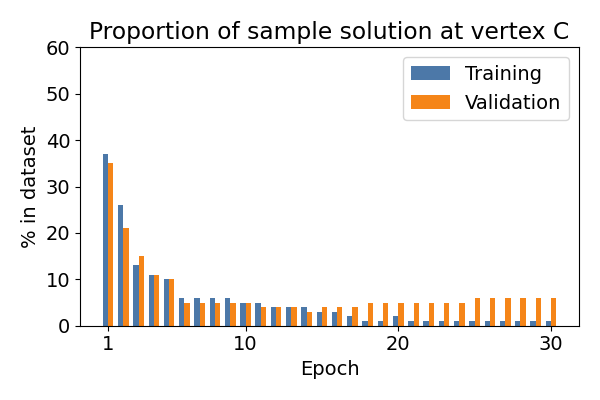}
    \caption{DPO model at seed 2}
  \end{subfigure}\hfill
  \begin{subfigure}{\imgw}
    \centering\includegraphics[width=\linewidth]{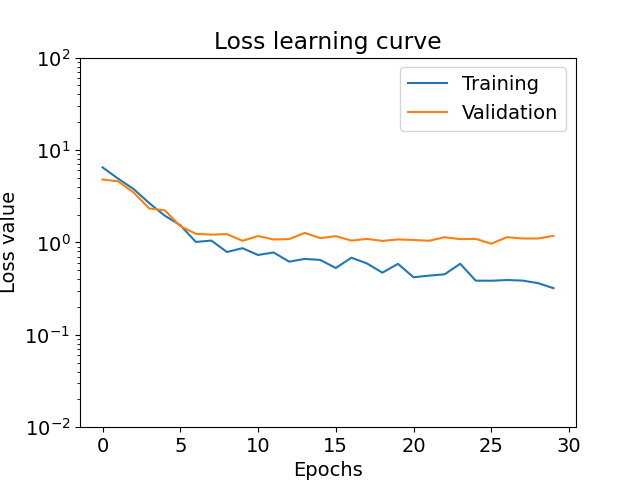}
    \caption{DPO model at seed 2}
  \end{subfigure}\hfill
  \par\vspace{0.5em}

  \begin{subfigure}{\imgw}
    \centering\includegraphics[width=\linewidth]{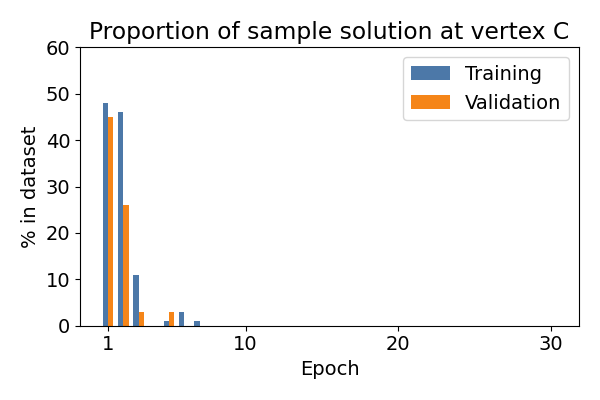}
        \caption{Regularized DPO model at seed 3}
  \end{subfigure}
  \begin{subfigure}{\imgw}
    \centering\includegraphics[width=\linewidth]{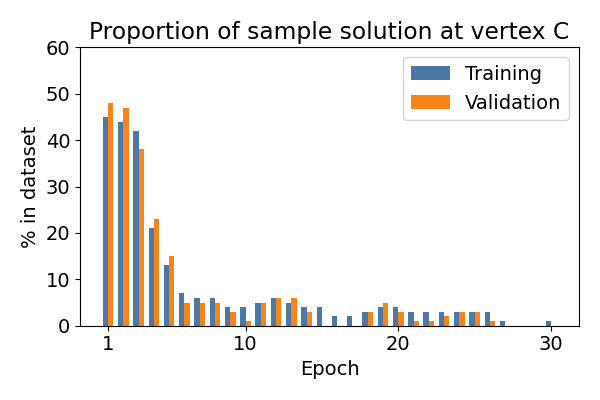}
    \caption{DPO model at seed 3}
  \end{subfigure}\hfill
  \begin{subfigure}{\imgw}
    \centering\includegraphics[width=\linewidth]{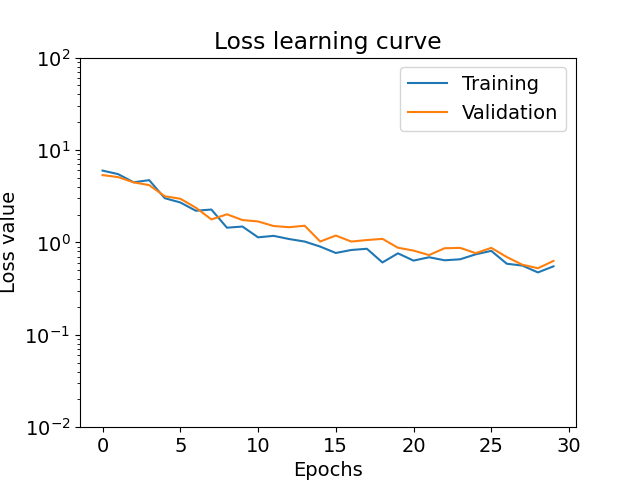}
    \caption{DPO model at seed 3}
  \end{subfigure}\hfill
  \par\vspace{0.5em}

  \begin{subfigure}{\imgw}
    \centering\includegraphics[width=\linewidth]{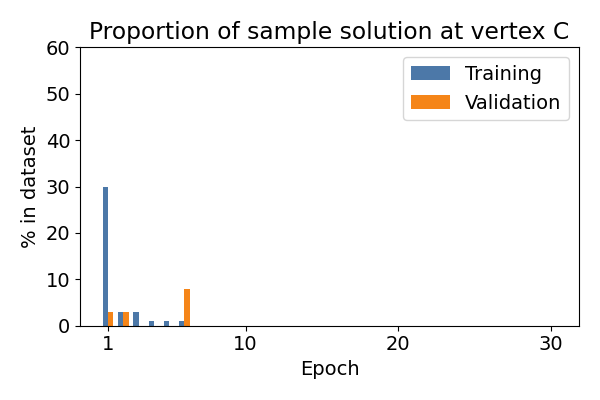}
        \caption{Regularized DPO model at seed 4}
  \end{subfigure}
  \begin{subfigure}{\imgw}
    \centering\includegraphics[width=\linewidth]{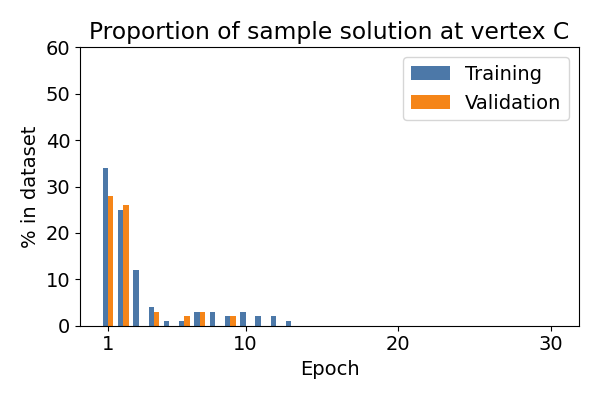}
    \caption{DPO model at seed 4}
  \end{subfigure}\hfill
  \begin{subfigure}{\imgw}
    \centering\includegraphics[width=\linewidth]{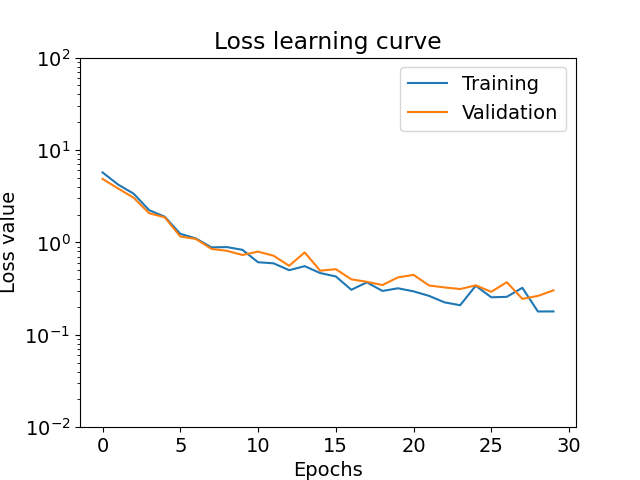}
    \caption{DPO model at seed 4}
  \end{subfigure}\hfill
  \par\vspace{0.5em}
  \caption{Problem category 1, seed 0 to 4: proportion of samples at vertex C and DPO model loss values.}
  \label{fig:te_2_1}
\end{figure}

\begin{figure}[!htbp]
  \centering

  % Each subfigure is half the text width minus some gutter
  \newcommand{\imgw}{0.32\linewidth}

  \begin{subfigure}{\imgw}
    \centering\includegraphics[width=\linewidth]{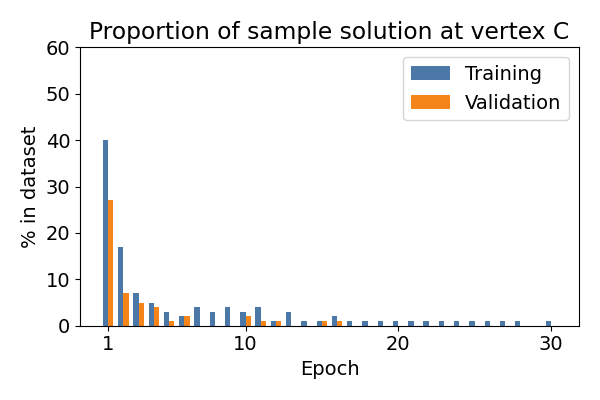}
    \caption{Regularized DPO model at seed 5}
  \end{subfigure}
\begin{subfigure}{\imgw}
    \centering\includegraphics[width=\linewidth]{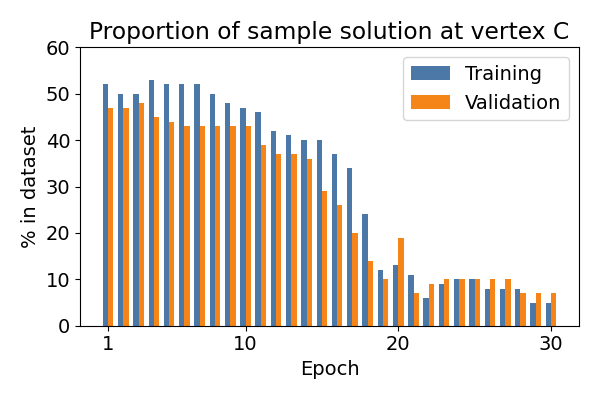}
    \caption{DPO model at seed 5}
  \end{subfigure}\hfill
  \begin{subfigure}{\imgw}
    \centering\includegraphics[width=\linewidth]{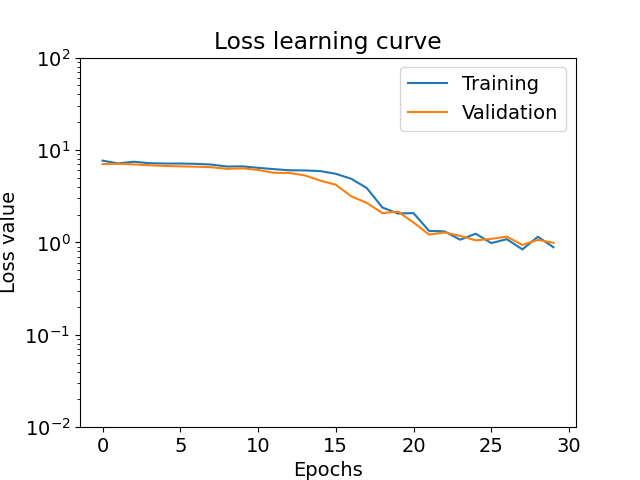}
    \caption{DPO model at seed 5}
  \end{subfigure}\hfill
  \par\vspace{0.5em}

  \begin{subfigure}{\imgw}
    \centering\includegraphics[width=\linewidth]{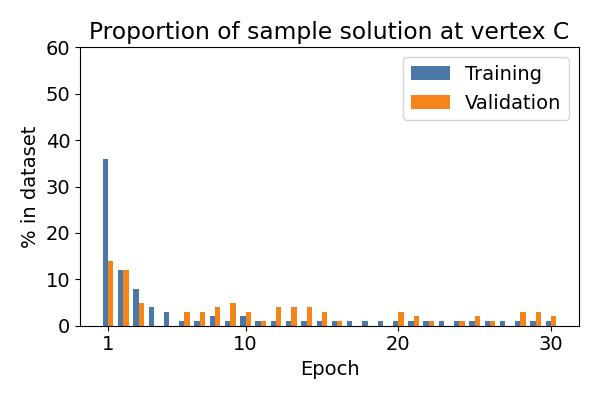}
        \caption{Regularized DPO model at seed 6}
  \end{subfigure}
  \begin{subfigure}{\imgw}
    \centering\includegraphics[width=\linewidth]{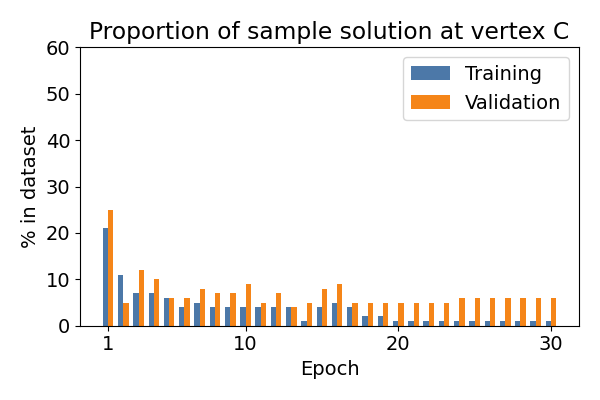}
    \caption{DPO model at seed 6}
  \end{subfigure}\hfill
  \begin{subfigure}{\imgw}
    \centering\includegraphics[width=\linewidth]{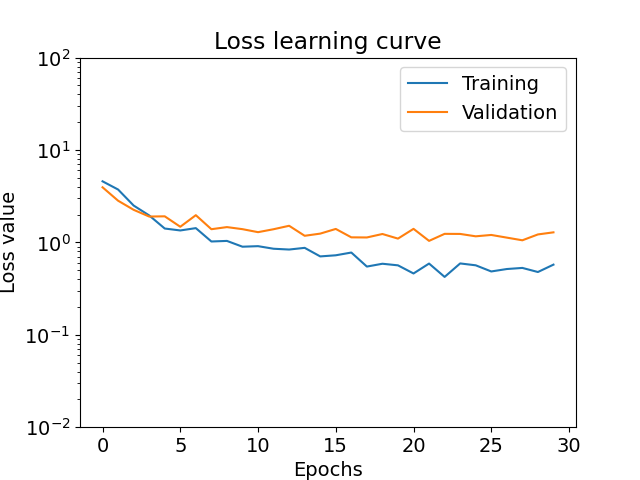}
    \caption{DPO model at seed 6}
  \end{subfigure}\hfill
  \par\vspace{0.5em}

  \begin{subfigure}{\imgw}
    \centering\includegraphics[width=\linewidth]{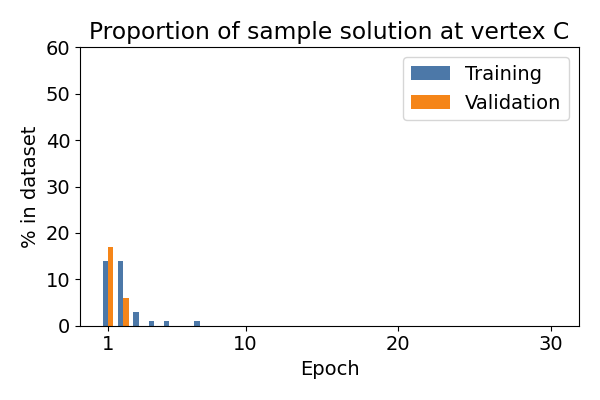}
        \caption{Regularized DPO model at seed 7}
  \end{subfigure}
  \begin{subfigure}{\imgw}
    \centering\includegraphics[width=\linewidth]{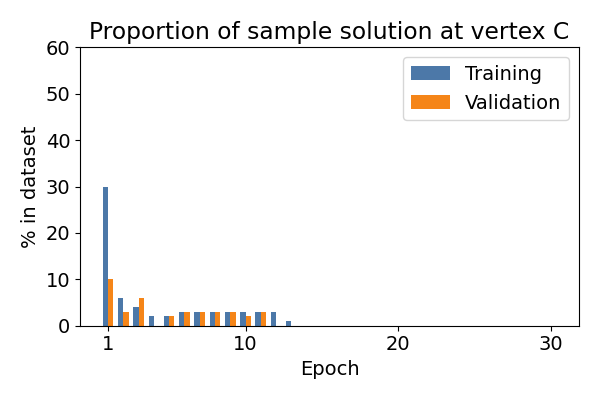}
    \caption{DPO model at seed 7}
  \end{subfigure}\hfill
  \begin{subfigure}{\imgw}
    \centering\includegraphics[width=\linewidth]{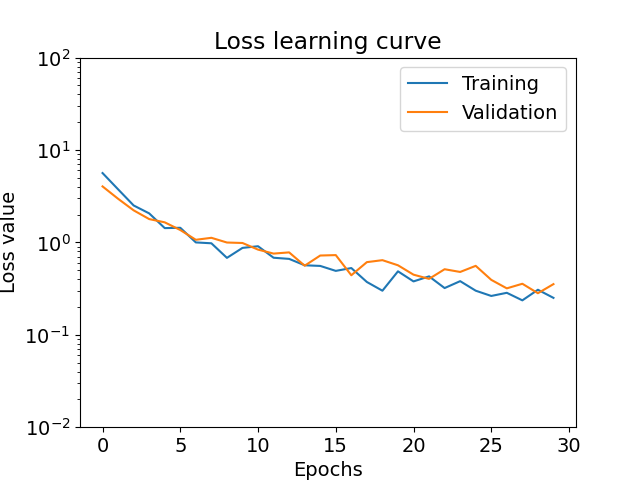}
    \caption{DPO model at seed 7}
  \end{subfigure}\hfill
  \par\vspace{0.5em}

  \begin{subfigure}{\imgw}
    \centering\includegraphics[width=\linewidth]{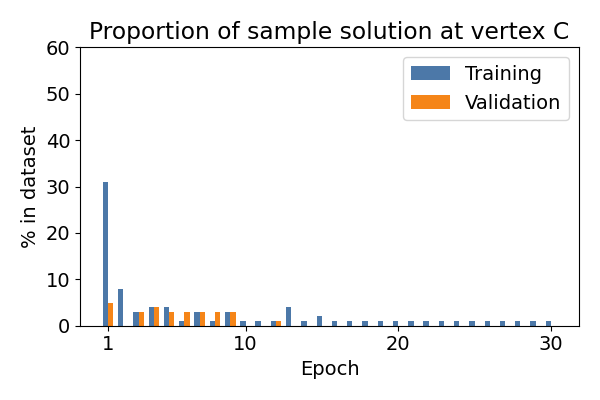}
        \caption{Regularized DPO model at seed 8}
  \end{subfigure}
  \begin{subfigure}{\imgw}
    \centering\includegraphics[width=\linewidth]{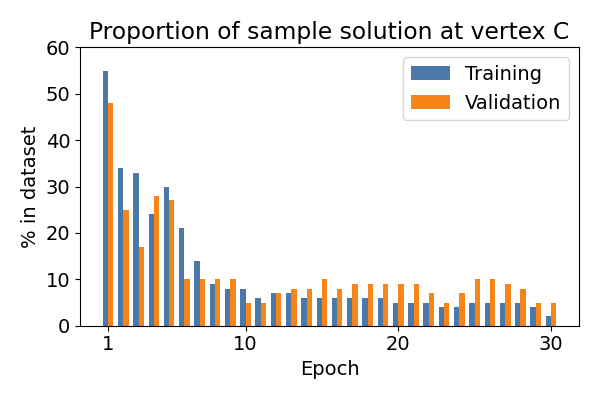}
    \caption{DPO model at seed 8}
  \end{subfigure}\hfill
  \begin{subfigure}{\imgw}
    \centering\includegraphics[width=\linewidth]{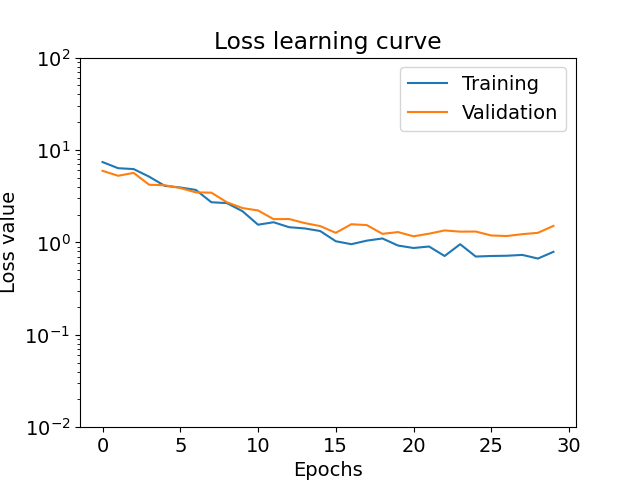}
    \caption{DPO model at seed 8}
  \end{subfigure}\hfill
  \par\vspace{0.5em}

  \begin{subfigure}{\imgw}
    \centering\includegraphics[width=\linewidth]{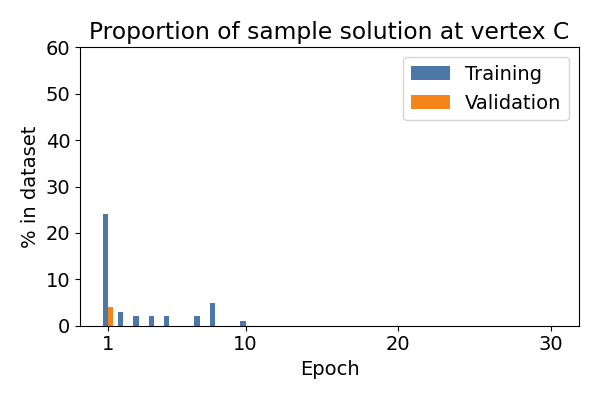}
        \caption{Regularized DPO model at seed 9}
  \end{subfigure}
  \begin{subfigure}{\imgw}
    \centering\includegraphics[width=\linewidth]{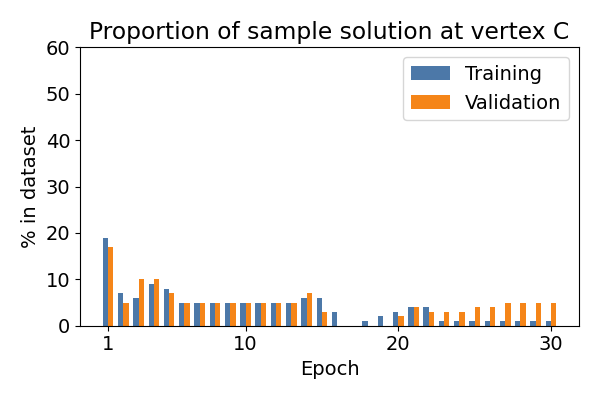}
    \caption{DPO model at seed 9}
  \end{subfigure}\hfill
  \begin{subfigure}{\imgw}
    \centering\includegraphics[width=\linewidth]{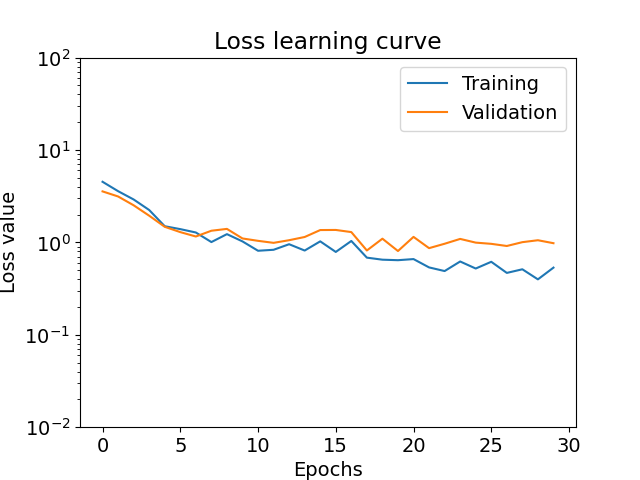}
    \caption{DPO model at seed 9}
  \end{subfigure}\hfill
  \par\vspace{0.5em}
  
  \caption{Problem category 1,  seed 5 to 9: proportion of samples at vertex C and DPO model loss values.}
  \label{fig:te_2_2}  \label{fig:te_1}
\end{figure}

\begin{figure}[!htbp]
  \centering

  % Each subfigure is half the text width minus some gutter
  \newcommand{\imgw}{0.32\linewidth}

  \begin{subfigure}{\imgw}
    \centering\includegraphics[width=\linewidth]{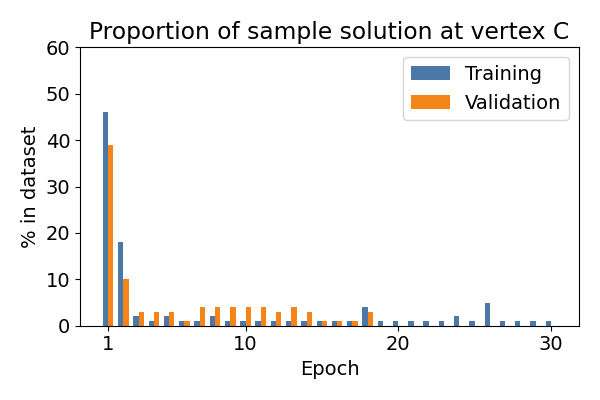}
    \caption{Regularized DPO model at seed 0}
  \end{subfigure}
  \begin{subfigure}{\imgw}
    \centering\includegraphics[width=\linewidth]{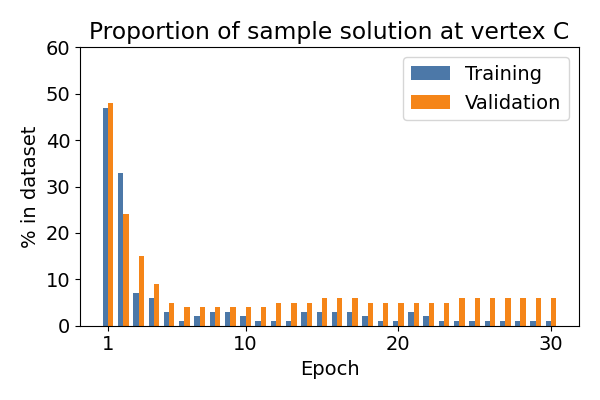}
    \caption{DPO model at seed 0}
  \end{subfigure}\hfill
  \begin{subfigure}{\imgw}
    \centering\includegraphics[width=\linewidth]{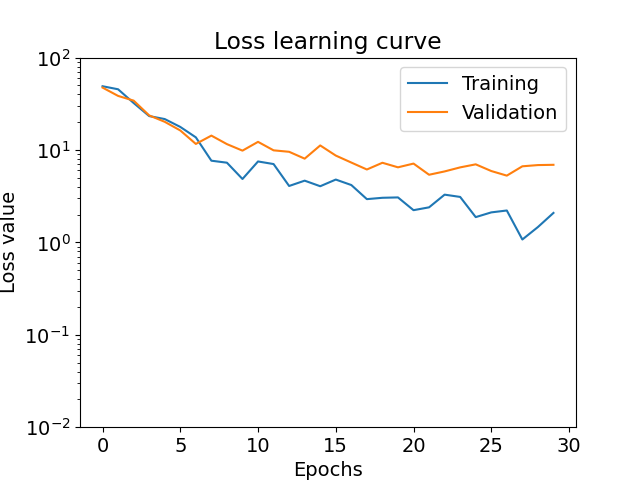}
    \caption{DPO model at seed 0}
  \end{subfigure}\hfill
  \par\vspace{0.5em}

  \begin{subfigure}{\imgw}
    \centering\includegraphics[width=\linewidth]{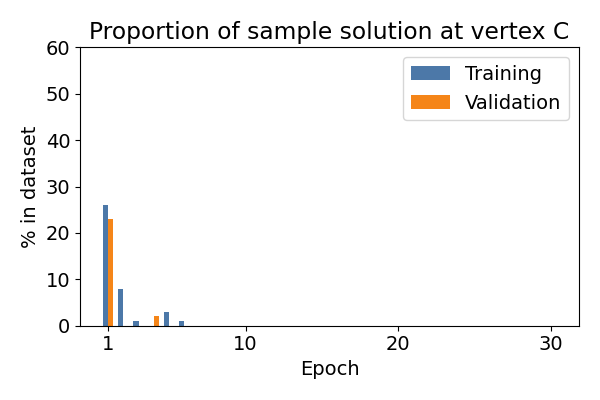}
        \caption{Regularized DPO model at seed 1}
  \end{subfigure}
  \begin{subfigure}{\imgw}
    \centering\includegraphics[width=\linewidth]{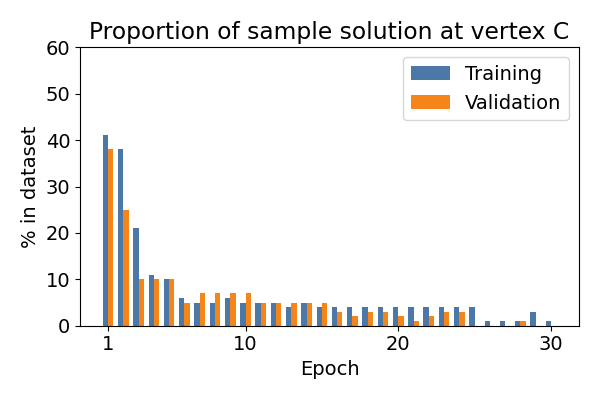}
    \caption{DPO model at seed 1}
  \end{subfigure}\hfill
  \begin{subfigure}{\imgw}
    \centering\includegraphics[width=\linewidth]{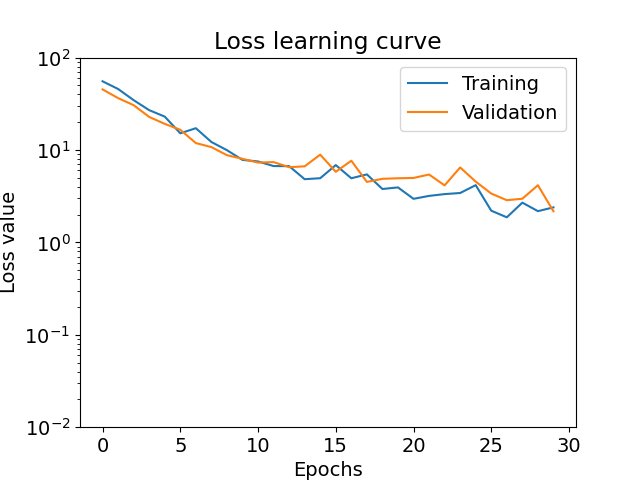}
    \caption{DPO model at seed 1}
  \end{subfigure}\hfill
  \par\vspace{0.5em}

  \begin{subfigure}{\imgw}
    \centering\includegraphics[width=\linewidth]{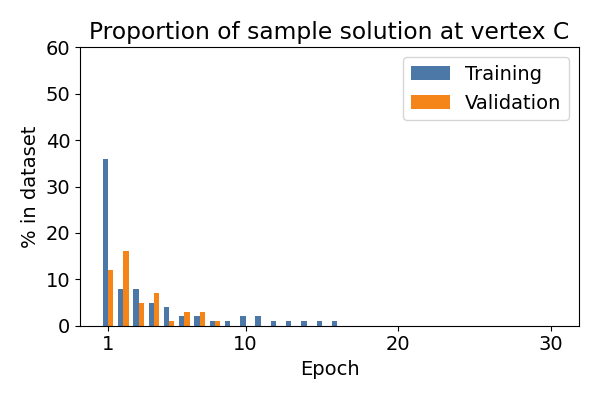}
        \caption{Regularized DPO model at seed 2}
  \end{subfigure}
  \begin{subfigure}{\imgw}
    \centering\includegraphics[width=\linewidth]{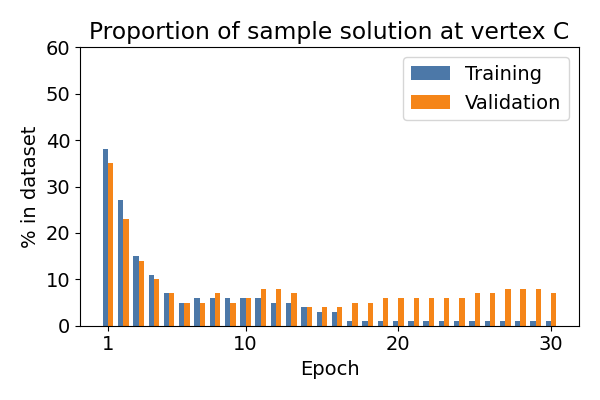}
    \caption{DPO model at seed 2}
  \end{subfigure}\hfill
  \begin{subfigure}{\imgw}
    \centering\includegraphics[width=\linewidth]{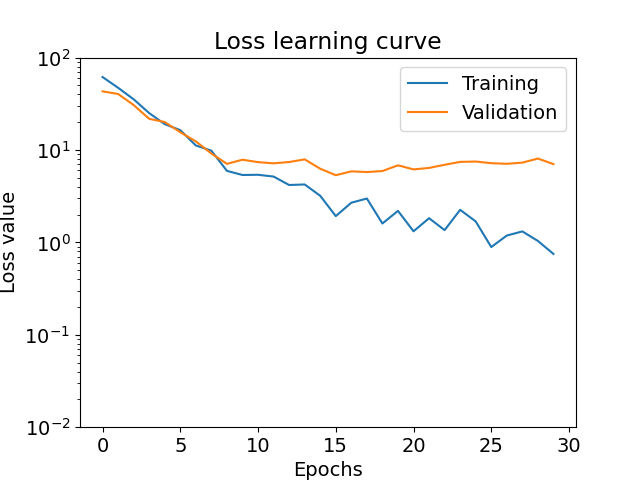}
    \caption{DPO model at seed 2}
  \end{subfigure}\hfill
  \par\vspace{0.5em}

  \begin{subfigure}{\imgw}
    \centering\includegraphics[width=\linewidth]{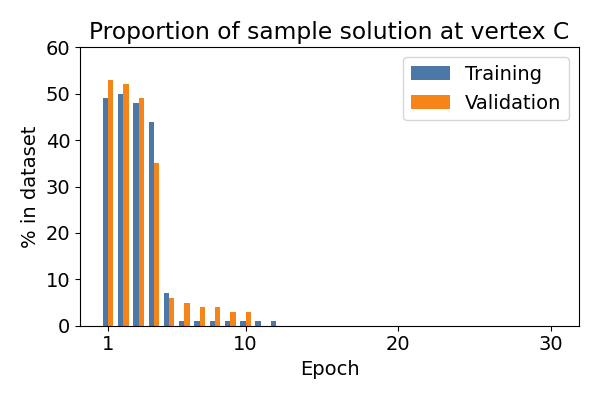}
        \caption{Regularized DPO model at seed 3}
  \end{subfigure}
  \begin{subfigure}{\imgw}
    \centering\includegraphics[width=\linewidth]{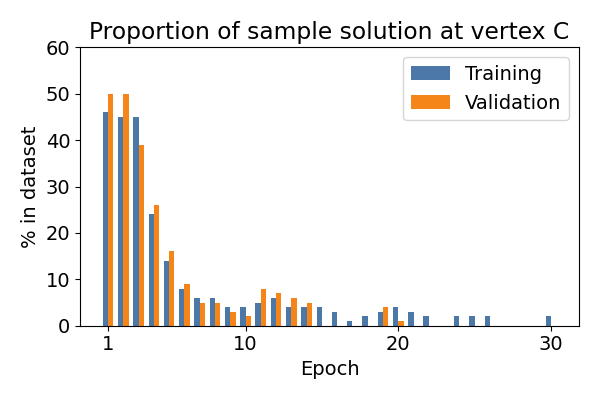}
    \caption{DPO model at seed 3}
  \end{subfigure}\hfill
  \begin{subfigure}{\imgw}
    \centering\includegraphics[width=\linewidth]{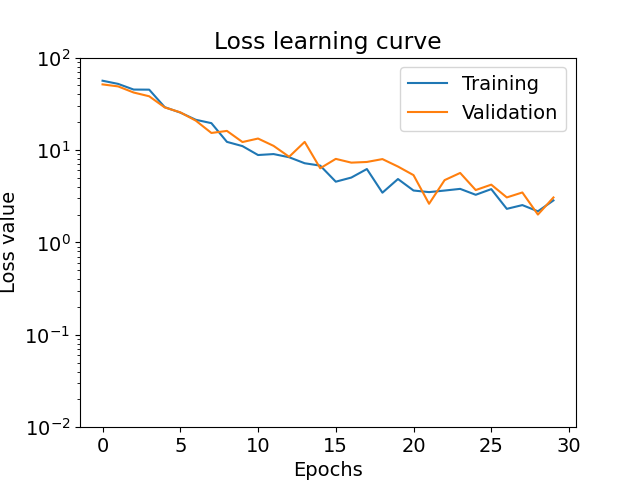}
    \caption{DPO model at seed 3}
  \end{subfigure}\hfill
  \par\vspace{0.5em}

  \begin{subfigure}{\imgw}
    \centering\includegraphics[width=\linewidth]{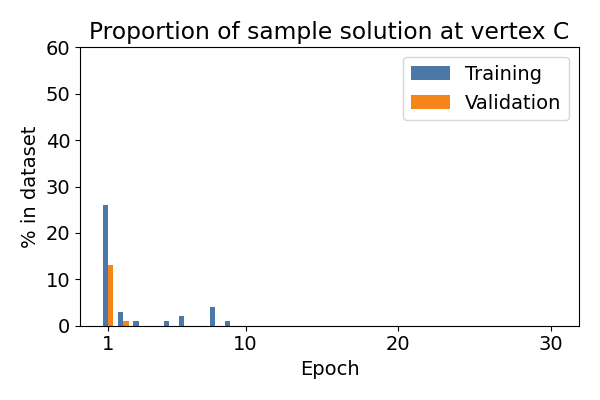}
        \caption{Regularized DPO model at seed 4}
  \end{subfigure}
  \begin{subfigure}{\imgw}
    \centering\includegraphics[width=\linewidth]{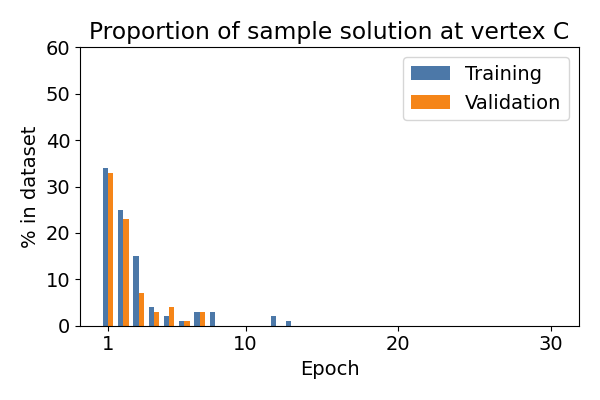}
    \caption{DPO model at seed 4}
  \end{subfigure}\hfill
  \begin{subfigure}{\imgw}
    \centering\includegraphics[width=\linewidth]{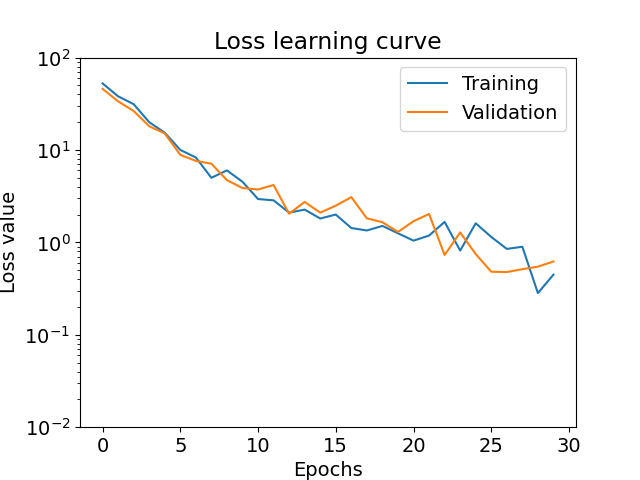}
    \caption{DPO model at seed 4}
  \end{subfigure}\hfill
  \par\vspace{0.5em}
  \caption{Problem category 2, seed 0 to 4: proportion of samples at vertex C and DPO model loss values.}
  \label{fig:te_3_1}
\end{figure}

\begin{figure}[!htbp]
  \centering

  % Each subfigure is half the text width minus some gutter
  \newcommand{\imgw}{0.32\linewidth}

  \begin{subfigure}{\imgw}
    \centering\includegraphics[width=\linewidth]{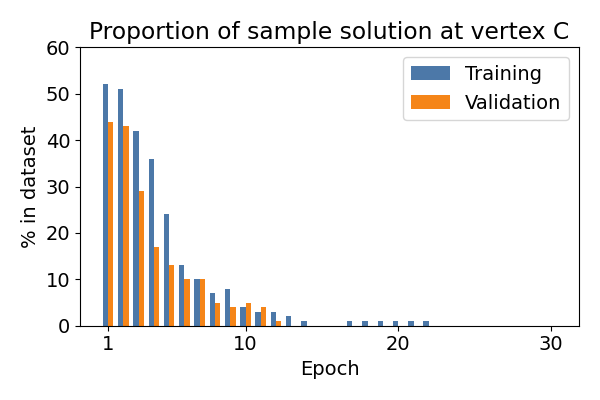}
    \caption{Regularized DPO model at seed 5}
  \end{subfigure}
\begin{subfigure}{\imgw}
    \centering\includegraphics[width=\linewidth]{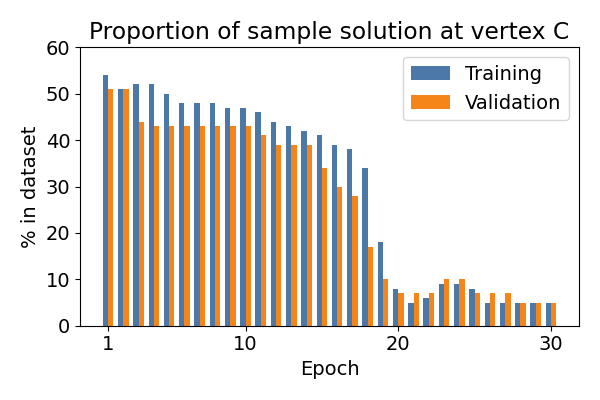}
    \caption{DPO model at seed 5}
  \end{subfigure}\hfill
  \begin{subfigure}{\imgw}
    \centering\includegraphics[width=\linewidth]{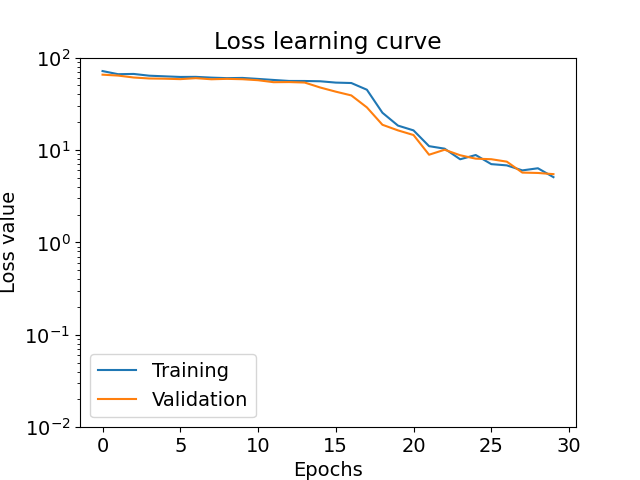}
    \caption{DPO model at seed 5}
  \end{subfigure}\hfill
  \par\vspace{0.5em}

  \begin{subfigure}{\imgw}
    \centering\includegraphics[width=\linewidth]{figure/DPO_reg_3_6_sample.png}
        \caption{Regularized DPO model at seed 6}
  \end{subfigure}
  \begin{subfigure}{\imgw}
    \centering\includegraphics[width=\linewidth]{figure/DPO_3_6_sample.png}
    \caption{DPO model at seed 6}
  \end{subfigure}\hfill
  \begin{subfigure}{\imgw}
    \centering\includegraphics[width=\linewidth]{figure/DPO_3_6_loss.png}
    \caption{DPO model at seed 6}
  \end{subfigure}\hfill
  \par\vspace{0.5em}

  \begin{subfigure}{\imgw}
    \centering\includegraphics[width=\linewidth]{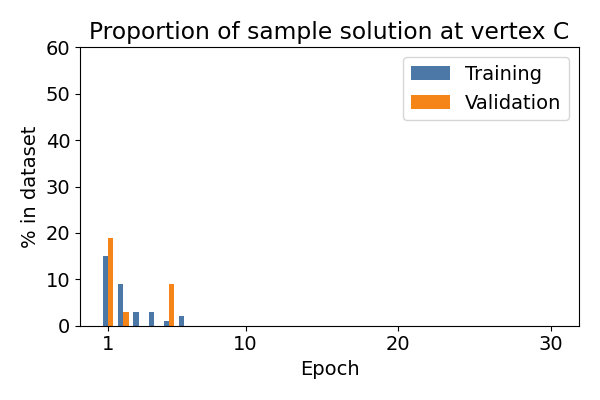}
        \caption{Regularized DPO model at seed 7}
  \end{subfigure}
  \begin{subfigure}{\imgw}
    \centering\includegraphics[width=\linewidth]{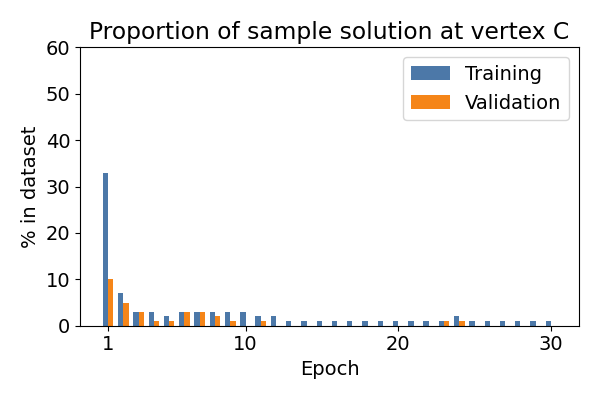}
    \caption{DPO model at seed 7}
  \end{subfigure}\hfill
  \begin{subfigure}{\imgw}
    \centering\includegraphics[width=\linewidth]{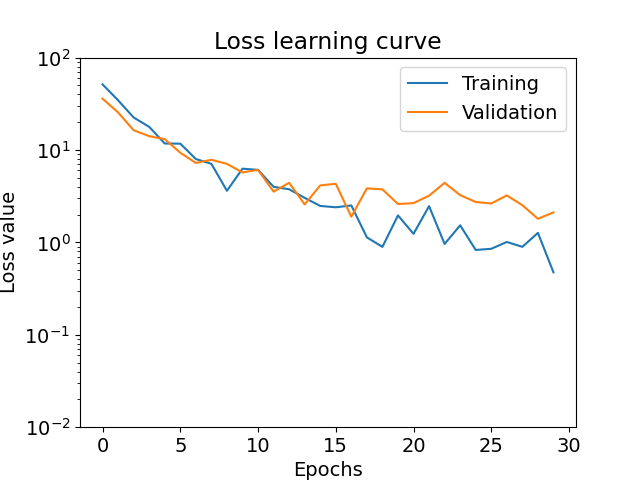}
    \caption{DPO model at seed 7}
  \end{subfigure}\hfill
  \par\vspace{0.5em}

  \begin{subfigure}{\imgw}
    \centering\includegraphics[width=\linewidth]{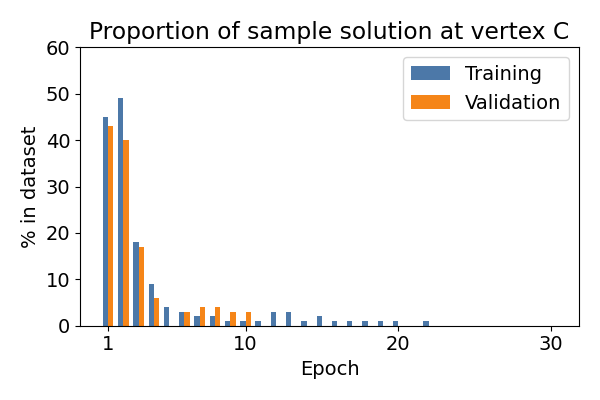}
        \caption{Regularized DPO model at seed 8}
  \end{subfigure}
  \begin{subfigure}{\imgw}
    \centering\includegraphics[width=\linewidth]{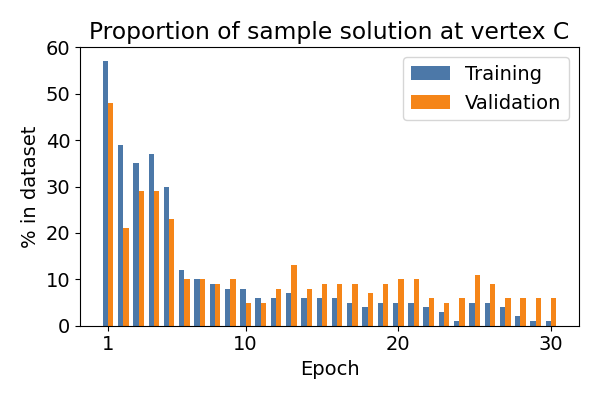}
    \caption{DPO model at seed 8}
  \end{subfigure}\hfill
  \begin{subfigure}{\imgw}
    \centering\includegraphics[width=\linewidth]{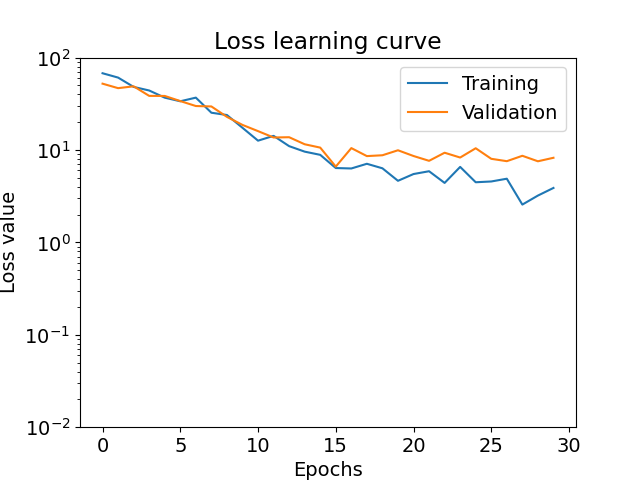}
    \caption{DPO model at seed 8}
  \end{subfigure}\hfill
  \par\vspace{0.5em}

  \begin{subfigure}{\imgw}
    \centering\includegraphics[width=\linewidth]{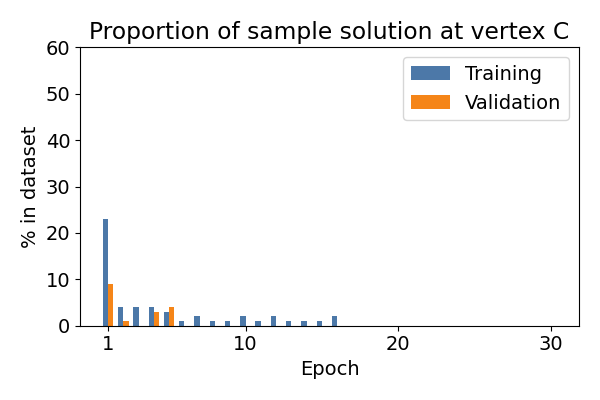}
        \caption{Regularized DPO model at seed 9}
  \end{subfigure}
  \begin{subfigure}{\imgw}
    \centering\includegraphics[width=\linewidth]{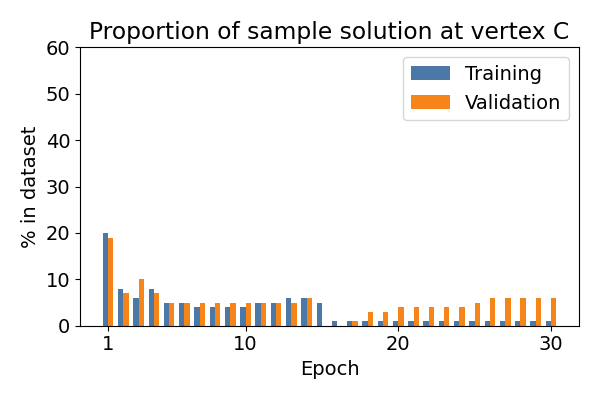}
    \caption{DPO model at seed 9}
  \end{subfigure}\hfill
  \begin{subfigure}{\imgw}
    \centering\includegraphics[width=\linewidth]{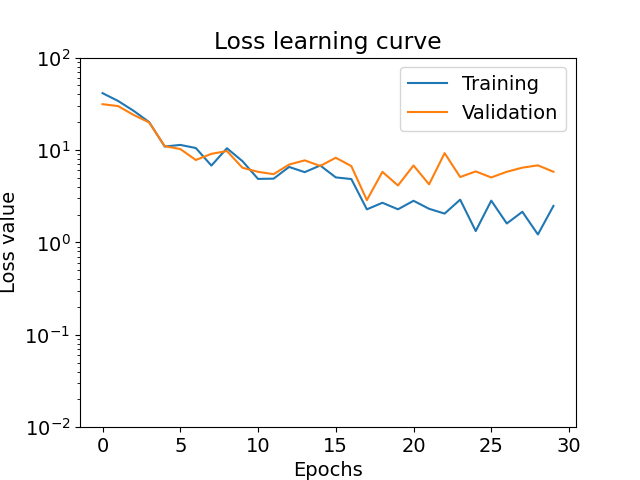}
    \caption{DPO model at seed 9}
  \end{subfigure}\hfill
  \par\vspace{0.5em}
  
  \caption{Problem category 2,  seed 5 to 9: proportion of samples at vertex C and DPO model loss values.}
  \label{fig:te_3_2} 
\end{figure}

We provide all figure regarding the specially designed toy problem studied in section \ref{sec:te}. Figures \ref{fig:te_2_1} to \ref{fig:te_3_2} display, for each problem category and each random seed, the proportion of samples whose optimal decision lies at vertex~C throughout training, computed on both the training and evaluation datasets and shown for both the standard and regularized variants. 

First, throughout learning, the proportion of optimal solutions at vertex $C$ in the training set is systematically lower for the regularized model than the standard one. Moreover, in case where that proportion in the training set decreases over the course of learning and eventually reaches zero, we observe that the corresponding proportion in the validation set generally stagnates at a strictly positive value.

We therefore deduce that the unregularized model fails to reliably learn to map cost vectors outside $\Theta_3$ and instead can only overfits the training samples exhibiting this pattern. This issue becomes increasingly pronounced in category 2, where decisions at vertex $C$ are even more detrimental and amplify model instability. By contrast, no such phenomenon is observed for the regularized model, which successfully mitigates these instabilities and thereby achieves superior performance. 

\end{document}